\documentclass[lettersize,journal]{IEEEtran}

\usepackage{amsmath,amsfonts,array,textcomp,stfloats,url,verbatim,graphicx,cite}

\usepackage{amssymb,bm,xcolor,nicefrac,stackengine,scalerel,soul, setspace}
\usepackage[ruled]{algorithm2e}
\usepackage[figuresright]{rotating}
\usepackage{booktabs}
\usepackage{multirow}
\usepackage[hidelinks]{hyperref}

\usepackage{threeparttable}

\newcommand{\trsp}{{\scriptscriptstyle\top}}

\newcommand{\diag}{\mathrm{diag}}

\newcommand{\kin}{{\scriptscriptstyle\text{kin}}}

\newcommand{\norm}[1]{\left\lVert#1\right\rVert}

\renewcommand{\figureautorefname}{Fig.}

\def\equationautorefname~#1\null{Eq. (#1)\null}
\def\figureautorefname~#1\null{Fig. #1\null}
\usepackage{tikz}
\usepackage{subfigure} 

\SetKwInput{KwData}{Input}
\SetKwInput{KwResult}{Output}

\begin{document}

\title{An Optimal Control Formulation of Tool Affordance Applied to Impact Tasks}

\author{Boyang Ti, Yongsheng Gao, Jie Zhao, Sylvain Calinon
     \thanks{This work was funded by the National Key Research and Development Program of China (No. 2022YFB4700701), and by the Major Research Plan of the National Natural Science Foundation of China (No. 92048301), supported by the China Scholarship Council, China (CSC, No. 202006120159). \emph{(Corresponding
authors: Yongsheng Gao; Sylvain Calinon.)}}
	\thanks{Boyang Ti is with the State Key Laboratory of Robotics and Systems,
		Harbin Institute of Technology, Harbin 150001, China, and also with Idiap Research Institute, CH-1920, Martigny, Switzerland (e-mail: 17b908043@stu.hit.edu.cn, tiboyang@outlook.com).}
	\thanks{Yongsheng Gao and Jie Zhao are with the State Key Laboratory of Robotics and Systems,
		Harbin Institute of Technology, Harbin 150001, China (e-mail: gaoys@hit.edu.cn, jzhao@hit.edu.cn).}
	\thanks{Sylvain Calinon is with Idiap Research Institute, 1920 Martigny, Switzerland, and also with Ecole Polytechnique Fédérale de Lausanne (EPFL), 1015 Lausanne, Switzerland (e-mail: sylvain.calinon@idiap.ch).}
}

\maketitle
    
\begin{abstract}
Humans use tools to complete impact-aware tasks such as hammering a nail or playing tennis. The postures adopted to use these tools can significantly influence the performance of these tasks, where the force or velocity of the hand holding a tool plays a crucial role. The underlying motion planning challenge consists of grabbing the tool in preparation for the use of this tool with an optimal body posture. Directional manipulability describes the dexterity of force and velocity in a joint configuration along a specific direction. In order to take directional manipulability and tool affordances into account, we apply an optimal control method combining iterative linear quadratic regulator (iLQR) with the alternating direction method of multipliers (ADMM). Our approach considers the notion of tool affordances to solve motion planning problems, by introducing a cost based on directional velocity manipulability. The proposed approach is applied to impact tasks in simulation and on a real 7-axis robot, specifically in a nail-hammering task with the assistance of a pilot hole. Our comparison study demonstrates the importance of maximizing directional manipulability in impact-aware tasks.
\end{abstract}

\section{Introduction}
\IEEEPARstart{E}{fficient} manipulation requires choosing a comfortable posture to execute an action, so that it is easy to generate a force or a velocity according to the tool affordances. The intrinsic of the above phenomenon is that the central nervous system (CNS) will transform information about the initial human posture and the task requirement into an appropriate pattern of muscular activity, where the target position is first transformed into a desired arm posture, which is then used to compute the motor commands \cite{bizzi1984posture, rosenbaum1995planning, desmurget1997postural}. 

The prevalent research in robot motion planning focuses on imitating the human hand motion, such as some typical point-to-point (pick-and-place) or trajectory following tasks (writing/wiping/polishing). However, as trajectories extend into a larger spatial area and/or task complexity increases, factors other than kinematics and biomechanics become increasingly relevant \cite{cos2011influence}. This demonstrates that arm morphology and its other intrinsic properties, such as dynamics, inertia, and muscle viscoelasticity, influence trajectory planning together with the environment geometry. Manipulability ellipsoid \cite{yoshikawa1984analysis, yoshikawa1985manipulability, yoshikawa1985dynamic} is taken as a geometric descriptor to measure and visualize the capability of a manipulator in a given configuration by considering the kinematic and dynamic information of the system. The manipulability projections along different directions provide insights regarding the directions along which the manipulator has higher or lower dexterity. A higher projection value means that a greater displacement of the end-effector can be produced in this direction, for the given joint angle configuration. Jaquier \emph{et al.} \cite{jaquier2018geometry} proposed a manipulability ellipsoid tracking method and validated it in a peg-in-hole task, where two kinds of desired manipulability are aligned referring to the central axis of the ellipsoids. In this article, we use directional manipulability \cite{chiu1988task} as a metric of desired manipulability, which can be flexibly transferred to a broader range of scenarios.

The concept of object/tool affordance originates from the field of psychology \cite{gibson1977theory}, whose definition has been extended to describe the relationship between the properties of an object and the capabilities of an agent that determine how the object could possibly be used \cite{norman1988psychology}. By extension, tool affordance refers to the qualities or characteristics of a tool that suggest the potential actions that can be performed with the tool. In this article, the notion of tool affordance will refer to the set of possible actions that could be generated according to the manner in which the tool is handled by the robot. It extends the function of the end-effector to enable the robot to expand its range of applications using tools, where the way in which the tool is grasped will determine in which manner the new end-effector can be used (e.g., using the tip of the hammer as the prolongation of the robot kinematic chain). Humans have the ability to generate skilled movements by exploiting tool affordances according to the demands of different tasks. It has been suggested that the body schema is plastic because it can incorporate external objects \cite{head1911sensory}. A hand-held tool, for example, may become so familiar to the user that it feels like a natural extension of the hand \cite{vaesen2012cognitive}. The assistance of external tools has dramatically increased the possibility of achieving more intractable tasks. Finding ways to manipulate the tools before reaching the final target pose is crucial. Therefore, uniting these two phases to complete motion planning is essential to transfer such skills to robots.

Fang \emph{et al}. \cite{fang2020learning} solved a tool-based manipulation as a two-stage problem consisting of a grasping phase and a manipulation phase, where the optimized task success is achieved by jointly training two networks. The dataset used in their network covers diverse shapes of hammers, allowing it to be generalized to many cases. However, there is still a gap between these two separate phases. Namely, how to transition between them naturally. In this article, we solve the motion planning problem by including the two phases, which allows the robot to anticipate in which manner the tool should be seized by the robot to be further used in an efficient way. Solving the optimization problem in such a way allows the viapoint separating the two phases and their transitions to be carefully chosen. In contrast, solving the problem in each phase separately would not provide the tool affordance capability of our approach. The \emph{tool affordance range} loosely refers to the desired range for the grasping of the tool that will determine how it can further be used in the hitting motion, which is specified as a viapoint constraint. The term \emph{viapoint} refers here to a key position that the robot needs to pass through during the movement. An optimal control problem (OCP) can be solved to generate optimal grasping and manipulation movements by balancing the weights of multiple cost terms, mimicking the human handling of the tool in different postures to facilitate the task according to the requirements. We use directional manipulability to describe the capability of using a tool. The maximization of the directional manipulability is taken as a part of the cost function to influence the tool grasping pose. To offer more choices to handle the tool, we represent the viapoint as a desired range treated as a spatial constraint, which allows the manipulator to choose any gripping point within the range while considering tool affordance. The tool handle gesture is adaptively optimized according to the demand of the subsequent manipulation task to be achieved.

An optimal control problem formulation can be used to solve this tool-use planning challenge. The iterative linear quadratic regulator (iLQR) \cite{mayne1966second, li2004iterative} is of particular interest to our work, as it can typically be used to search for a solution in the joint angle configuration space of the robot. To consider the inequality constraints that occur in the task space introduced by the tool affordance, there are several solvers available for solving constrained optimization problems, such as SNOPT \cite{gill2005snopt}, SLSQP \cite{kraft1988software} and IPOPT \cite{wachter2002interior}. The optimization problems in the field of robotics manipulation can be solved efficiently by the above solution methods. The alternating direction method of multipliers (ADMM) \cite{boyd2011distributed} is another popular optimization technique for solving problems that can be decomposed into subproblems with different variables. In this article, inspired by Sindhwani \emph{et al}. work \cite{sindhwani2017sequential} on car parking and obstacle avoidance, we embed ADMM into iLQR to solve a constrained motion planning problem and demonstrate its effectiveness in serial manipulator systems through point-to-range and pick-and-place tasks. 
The ADMM-iLQR approach that we use is similar to the method proposed in \cite{ma2022alternating}, except that we employ an approximation for the optimization function to make it approximately equal to the convex problem in the first ADMM iteration. In \cite{ma2022alternating}, the control case of iLQR is considered, which can be solved by dynamic programming. In our work, the planning case is instead considered, where the solution can be obtained analytically through the batch solution of iLQR, which can efficiently computed in matrix form by only relying on linear algebra. Note that in both control and planning cases, the underlying iterative approach is not guaranteed to lead to a global optimum. However, the local optimal solution that it finds remains an appropriate estimate in a wide range of problems in robotics, including redundant problems with sparse costs in the forms of viapoints as the one that we target in this article.

\begin{figure*}
	\centering
	\includegraphics[width=0.7\textwidth]{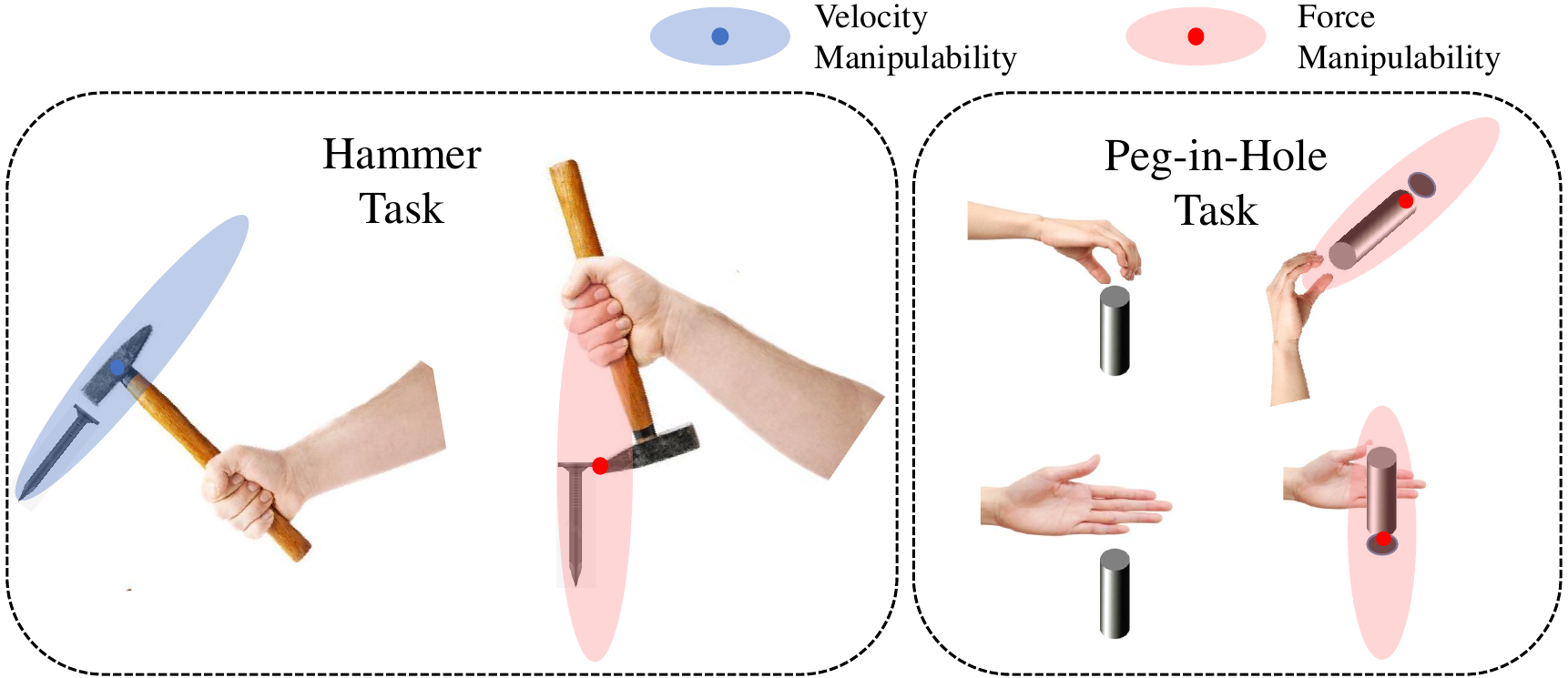}	
	\caption{\emph{Left:} The ways in which tools are seized and used vary according to the characteristics of the task (tool affordances). For hammering a nail, we grasp the hammer in a way that is efficient for the task demands, since the greater the velocity, the deeper the nail can be driven. For nail-pulling, the same hammer is grasped in a different way. Here, the greater the static force applied to the nail, the easier the nail can be pulled out. 
    \emph{Right:} In a peg-in-hole task, we typically choose a comfortable pose to insert a peg into a narrow hole, which depends on the relations between the hand, the peg and the hole (with maximum force manipulability along the insertion direction).}
 	\label{Fig2}
	\vspace{-0.3cm}
\end{figure*}

From the perspective of biomechanics, we can observe the characteristics of human behaviors in impact-aware tasks with tool assistance. For example, as shown in \autoref{Fig2}, when we select a grasp to seize an object, we consider the relationship among body configuration, tool, and target poses. When we use a claw hammer to pull out a nail driven into a wooden board, we typically prefer to hold the arm in a tucked-in configuration, making it easier to generate substantial static forces between the nail and the claw. In contrast, when we want to drive a nail into a board in a more stretched configuration, we can easily drive the arm to a higher speed and transfer it to the hammer, as a way to generate maximum momentum (provided that the inertia of the hammer remains constant). In addition to high speeds, the inertial and stiffness properties of the arm configuration play a critical role in such impact-aware tasks, where we need our muscles to absorb the vibrations while controlling the tool's stability during the impact. In the setup that we will consider, instead of absorbing vibrations solely through the properties of the arm, a pilot hole is additionally used to prevent deviations arising from uneven resistance distribution at the tip of the nail. This external assistance guarantees a smooth execution of the hammering task, which is also widely applied in our daily lives.


The contributions of this article can be summarized as follows:

(1) We employ an ADMM-iLQR strategy to solve an optimal control problem (OCP) considering tool affordance constraints in impact-aware tasks;

(2) We introduce a maximum directional manipulability cost in the optimal control framework to optimize the grasping and final manipulation posture;

(3) We make a comprehensive comparison of different approaches to measuring manipulability in directional tasks;

To fully illustrate the significance of these contributions, we provide simulations to validate our approach and demonstrate its application in an impact-aware task (hammering).
   
Compared to the existing constrained OCP, the use of ADMM solver in our work is employed to include a tool affordance aspect, specified as a range that can be used during robot manipulation. We show that this convex constraint can be used to enlarge the manipulation capability of the robot. The inequality constraint problem is often studied in the context of obstacle avoidance or joint physical constraint avoidance. We show here that it can also be used as a simple model of tool affordance described as a constraint in OCP.

The rest of the paper is organized as follows. In Section \ref{sec:related work}, we summarize the related works on impact-aware hammering and tool-use tasks, constrained optimal control and manipulability. In Section \ref{sec:Preliminaries}, we introduce the background knowledge related to our method. In Section \ref{sec:Definitions}, we introduce our proposed method by deriving the control policy with constraints and analyze the directional manipulability and other soft constraints considered in the cost. In Section \ref{sec:simulation} and \ref{sec:robot experiment}, we compare our approach with methods in which the manipulability is not taken into account, methods using manipulability determinant metrics \cite{yoshikawa1985manipulability}, and method minimizing manipulability distances \cite{jaquier2018geometry}. In Section \ref{sec:discussion} and \ref{sec:conclusion}, we discuss current limitations and future work.

\section{Related Work}
\label{sec:related work}
\subsection{Impact-Aware Hammering Task and Tool-Use Task}
\label{subsec: impact aware task}
Hammering is a typical impact-aware task, common in daily life, industry, and even in the medical field. The general performance index of hammering is the speed of the hammer head in contact with the object or landmark of interest \cite{izumi1997hitting}. Research on hammering has been validated with single joint \cite{garabini2011optimality, jujjavarapu2023variable}, serial manipulators \cite{romanyuk2022multiple} and humanoid robots \cite{tsujita2008analysis, matsumoto2006humanoid}. Tsujita \emph{et al.} \cite{tsujita2008analysis} presented a model of contact dynamic and proposed an evaluation of the impulsive force prediction model. Imran \emph{et al.} \cite{imran2017closed} focused on minimizing the impulses on robot joints to increase their lifespan. Jujjavarapu \emph{et al.} \cite{jujjavarapu2023variable} demonstrated the usability of optimization for exploiting the dynamics of a variable stiffness mechanism (VSM) in hammering tasks to improve task performance with time-varying stiffness profiles. The aspect of vibration absorbing or post-impact system stability has been well investigated in past decades. Fang \emph{et al.} \cite{fang2020learning} and Tanev \emph{et al.} \cite{tanev2022joint} developed a model that learns policies for task-oriented grasping and task-related tool manipulation with self-supervision. Training neural networks takes time, requires large datasets, and produces results that lack interpretability. It also rarely takes into account the known biomechanical aspects. In our work, we consider the pose configuration for the pre-hammering state, which is achieved by using an optimal control method. This allows the manipulator to limit the control commands while generating important hammering speed in the desired configuration.

A few works have used a learning from demonstration strategy to achieve tool-based manipulation skills learning. Li \emph{et al.} \cite{li2015teaching} proposed a hierarchical architecture to embed the use of tools in a learning from demonstration framework. Rajeswaran \emph{et al.} \cite{rajeswaran2017learning} incorporated human demonstrations in deep reinforcement learning to provide robust manipulation strategies. In addition to offline learning, Jain \emph{et al.} \cite{jain2014learning} proposed an online and incremental approach to make the robot acquire manipulation capabilities by interaction with the human user. Involving humans in the learning process is a very efficient way to exploit the redundancy of robot arms, but it can be challenging to imitate human manipulation postures just through kinesthetic or visual demonstrations, especially when the demonstrated skill requires to be close to the robot limits. Kunz \emph{et al.} \cite{kunz2014probabilistically} presented an RRT motion planner that considers joint acceleration limits and potentially non-zero start and goal velocities in hammering tasks. Holladay \emph{et al.} \cite{holladay2019force} formulated kinematic and force limits as decision variables with various constraints. In this article, we discuss the influence of the impact-aware task requirements on the tool grasping posture and consider this influence in an optimal control formulation of motion planning.

\subsection{Constrained Optimal Control}
For real-world application scenarios, different constraints should be considered in the optimization problem, such as dynamic constraints, task constraints (incl. obstacle avoidance), and physical limit constraints (incl. robot joint position and velocity limits). 
Control-limited differential dynamic programming \cite{tassa2014control} can be used to handle box inequality constraints at the control level. Chen \emph{et al.} \cite{chen2017constrained} proposed a constrained iLQR approach that transforms the constraints to a cost function by adding a barrier function, which can also be extended to a logarithmic barrier \cite{chen2019autonomous}.

Nowadays, the development of numerical optimization has made unprecedented progress. ADMM has been proposed to solve optimal control problems by showcasing excellent efficiency and stability. It has been used in diverse areas such as autonomous driving \cite{ma2022alternating}, electricity networks \cite{braun2018hierarchical} or wind farms \cite{liao2021distributed}. It can handle problems with multiple constraints and it can be more efficient than the distributed gradient method\cite{ling2014decentralized}, especially in problems in which the objective function can be separated into two or more subproblems that can be solved independently. Hong \emph{et al.} \cite{hong2017linear} established the global (linear) convergence of the ADMM method for a class of convex objective functions involving any number of blocks.  In some special conditions (strong convexity and Lipschitz differentiable), variants of ADMM can ensure linear convergence. Benefiting from the subproblems split strategy, the ADMM algorithm can relieve the computation burden caused by increasing the system dimensions. Also, ADMM is typically less dependent on parameter settings than many other distributed methods and, therefore, is easy to implement. Overall, ADMM can often offer performance comparable to specialized algorithms, and in most cases, the simple ADMM algorithm will be efficient enough to be useful in many robotics problems \cite{boyd2011distributed}, see Yang \emph{et al.} \cite{yang2022survey} for a comprehensive survey on the fundamental property and optimization options in ADMM. In robot manipulation tasks, it has also been used in contact-rich optimization problems. Aydinoglu \emph{et al.} \cite{aydinoglu2022real, aydinoglu2023consensus} proposed using ADMM to exploit its distribution to solve multi-contact dynamics and complicated contact-rich manipulation tasks. Shirai \emph{et al.} \cite{shirai2022simultaneous} proposed a distributed optimization framework based on ADMM to solve contact dynamics. Shorinwa \emph{et al}. \cite{shorinwa2021distributed} used the ADMM method to solve a contact-implicit trajectory optimization problem in a multi-agent system. Wijayarathne \emph{et al.} \cite{wijayarathne2023real} employed ADMM to generate real-time optimal control in soft contact problems. The above collection of work focused on contact-rich optimization problems. Our work exploits the power of distributed fast solvers to handle constrained optimization problems encapsulating manipulability and tool affordance information, that we exploit in impact-aware tasks. This problem is critical for ensuring tool affordance constraints and optimizing the manipulation configuration of the robotic manipulator to produce optimal impact actions. 

\subsection{Manipulability Ellipsoids}
Velocity and force manipulability ellipsoids are two descriptors used to measure the ability of an end-effector to perform velocity or force along all task-space directions for a given joint angle configuration. A typical metric based on these descriptors consists of keeping only the volume of the ellipsoid as the source of information \cite{yoshikawa1985manipulability}. Mari{\'c} \emph{et al.} \cite{Maric21} used a Riemannian metric to provide a measure of proximity to singular configurations. A thorough study of manipulability ellipsoid on learning, tracking, and transfer has been conducted by Jaquier \emph{et al.} \cite{jaquier2021geometry, jaquier2018geometry, rozo2017learning}, where a control formulation was developed to track desired manipulability ellipsoids, which can be extended to nullspace control to also reach a target pose as primary goal. Here, we use the ADMM-iLQR method to extend the approach to viapoint tasks characterized by a desired range to reach, which is used to pick and manipulate a hammer, by letting the robot exploits the tool affordances to determine optimal picking locations by anticipating the next part of the task that consists of hammering a nail.

In this article, we propose to exploit directional manipulability in the cost function, defined as the length of the manipulability ellipsoid along a particular direction of interest. 
Tugal \emph{et al.} \cite{tugal2022manipulation} implement directional manipulability to extend the interaction capabilities of a mobile manipulator on the valve turning task. Kim \emph{et al.} \cite{kim2021evaluating} evaluated the maximum directional kinematic capability using optimization-based methods for redundant manipulators. Besides improving the velocity and force, Marais \emph{et al.} \cite{marais2021anisotropic} exploited this measure to maximize the dynamic manipulability to reject directional disturbances. For the hammering task, the direction of interest corresponds to the orientation of the nail perpendicular to the board. We incorporate this into the cost function to account for its effect on the grasping posture.

\section{Preliminaries}
\label{sec:Preliminaries}

\subsection{Alternating Direction Method of Multipliers (ADMM)}

%
Alternating Direction Method of Multipliers (ADMM) is an algorithm designed to combine the decomposability of dual ascent methods with the superior convergence properties of the method of multipliers \cite{boyd2011distributed}. The algorithm solves problems in the form
\begin{equation}
	\begin{array}{rl}
		\underset{\bm{x},\bm{z}}{\min} \quad c(\bm{x})+g(\bm{z}) \\
		\text{ s.t. } \quad \bm{Ax}+\bm{Bz}=\bm{d},
	\end{array}
    \label{eq:admm_consensus}
\end{equation}
with variable $\bm{x}\in \mathbb{R}^n$ and $\bm{z}\in{\mathbb{R}}^m$, where $\bm{A}\in\mathbb{R}^{p\times n}, \bm{B}\in\mathbb{R}^{p\times m}$ and $\bm{d}\in\mathbb{R}^p$. $c(\cdot)$ and $g(\cdot)$ are assumed to be convex. The difference from the general linear equality-constrained problem is that the variable has been split into two parts, called $\bm{x}$ and $\bm{z}$ here, with the objective function separable across this splitting. The augmented Lagrangian function of this is
\begin{equation}
	\begin{array}{rl}
			\bm{L}_\rho(\bm{x}, \bm{z}, \bm{y}) = & c(\bm{x}) + g(\bm{z}) + \bm{y}^\trsp(\bm{Ax}+\bm{Bz}-\bm{d})\\
		&+(\rho/2)\norm{\bm{Ax}+\bm{Bz}-\bm{d}}^2_2,
	\end{array}
\end{equation}
where $\rho > 0$ is a penalty parameter and $\bm{y}$ is the dual variable. The formulation can be described compactly by combining the linear and quadratic terms in the augmented Lagrangian and scaling the dual variable. The scaled dual variable is defined as $\bm{\lambda}=(1/\rho)\bm{y}$. Then, by following \cite{boyd2011distributed}, the iteration process of ADMM becomes 
\begin{align}
		\bm{x}^{k+1} &= 
  \arg\underset{\bm{x}}{\min} \left(c(\bm{x})+\frac{\rho}{2}\norm{\bm{Ax}+\bm{B}\bm{z}^{k}-\bm{d}+\bm{\lambda}^{k}}^2_2\right), \nonumber\\
		\bm{z}^{k+1} &= 
  \arg\underset{\bm{z}}{\min}  \left(g(\bm{z})+\frac{\rho}{2}\norm{\bm{A}\bm{x}^{k+1}+\bm{B}\bm{z}-\bm{d}+\bm{\lambda}^{k}}^2_2\right), \nonumber\\
		\bm{\lambda}^{k+1} &= \bm{\lambda}^k + \bm{Ax}^{k+1}+\bm{Bz}^{k+1}-\bm{d}.
     \label{eq:admm_scaled}
\end{align}

The convergence condition of ADMM solver is divided into:
\begin{enumerate}
	\item[-] Residual convergence: $\bm{Ax}^{k+1}+\bm{Bz}^{k+1}-\bm{d} \rightarrow0$
	\item[-] Objective convergence: $c(\bm{x}^{k+1})+g(\bm{z}^{k+1})$ approaches the optimal value 
	\item[-] Scaled dual variable convergence: $\bm{\lambda} \rightarrow \bm{\lambda}^*$ , where $\bm{\lambda}^*$ is a dual optimal variable
\end{enumerate}

Consider the following generic constrained convex optimization problem over the variable $\bm{x}$
\begin{equation}
	\begin{array}{rl}
			\underset{\bm{x}}{\min} & c(\bm{x}) \\
			\text { s.t. } & \bm{x} \in \mathcal{C},
		\end{array}
  \label{eq:ADMM_generic}
\end{equation}
where $c(\cdot)$ and $\mathcal{C}$ are convex. This problem can be written in an ADMM consensus formulation as 
\begin{equation}
	\begin{array}{rl}
		\underset{\bm{x}}{\min} \quad c(\bm{x})+g(\bm{z}) \\
		\text{ s.t. } \quad \bm{x}-\bm{z}=\bm{0},
	\end{array}
    \label{eq:admm_consensus}
\end{equation}
where $g$ is the indicator function of $\mathcal{C}$, \emph{i.e.}, $g(\bm{z})=0$ for $\bm{z} \in \mathcal{C}$ and $g(\bm{z})=+\infty$ otherwise, and $\bm{z}$ is related to the constraint $\mathcal{C}$.
The augmented Lagrangian becomes
\begin{equation}
	\begin{array}{rl}
			\bm{L}_\rho(\bm{x}, \bm{z}, \bm{\lambda}) = & c(\bm{x}) + g(\bm{z})+(\rho/2)\norm{\bm{x}-\bm{z}+\bm{\lambda}}^2_2.
	\end{array}
 \label{eq:admm_convex}
\end{equation}

Therefore, the scaled form of ADMM for this problem is
\begin{equation}
	\begin{aligned}
			\bm{x}^{k+1} &= \arg\underset{\bm{x}}{\min }\left(c(\bm{x})+\rho \left\|\bm{x}-\bm{z}^k+\bm{\lambda}^k\right\|_2^2\right), \\
			\bm{z}^{k+1} & =\Pi_{\mathcal{C}}\left(\bm{x}^{k+1}+\bm{\lambda}^k\right), \\
			\bm{\lambda}^{k+1} & =\bm{\lambda}^k+\bm{x}^{k+1}-\bm{z}^{k+1}.
		\end{aligned}
\end{equation}

The $\bm{x}$ update involves minimizing $c(\cdot)$ plus a convex quadratic function. The $\bm{z}$ update is the Euclidean projection onto $\mathcal{C}$.
In the Appendix, we show that in the case of affine constraints, the projection step corresponds to the projection onto the sublevel set of a convex function. 

\subsection{Manipulability Ellipsoids}

The velocity and force manipulability ellipsoids are two mutually complementary metrics, that can be obtained by an inverse transformation of the other. Here, we consider velocity manipulability, which describes the volume that can be reached by the end-effector in Cartesian space for a single bounded command in joint space. It will typically describe how the robot will be able to face perturbations at a given time step, which is an important descriptor for kinematically redundant robots as it describes how the selected robot posture can influence the movement of the end-effector in all directions of the task space.

The velocity manipulability of a robot can be found by using the kinematic relationship between task velocity $\bm{\dot{x}}$ and joint velocity $\bm{\dot{q}}$
\begin{equation}
	\bm{\dot{x}} = \bm{J}(\bm{q})\bm{\dot{q}},
	\label{eq:base_kine}
\end{equation}
where $\bm{q}\in \mathbb{R}^n$ and $\bm{J}\in \mathbb{R}^{6 \times n}$ are the joint position and Jacobian of the forward kinematics function describing the robot, respectively. The derivation of the manipulability ellipsoid typically starts by taking the joint velocity limited in a unit sphere by
\begin{equation}
	{\Vert \bm{\dot{q}} \Vert} = 1,
\end{equation}
therefore we can obtain
\begin{equation}
	\bm{\dot{q}}^\trsp \bm{\dot{q}}=\bm{\dot{x}}^\trsp(\bm{J}\bm{J}^\trsp)^{-1}\bm{\dot{x}}=1,
\end{equation}
by using the least-squares inverse kinematics relation $\bm{\dot{q}}=\bm{J}^{\dagger}\bm{\dot{x}}=\bm{J}^\trsp(\bm{J}\bm{J}^\trsp)^{-1}\bm{\dot{x}}$. We can observe that under this constraint, the matrix $\bm{J}\bm{J}^\trsp$ determines the scalability of the velocity manipulability described by 
\begin{equation}
	\bm{M}(\bm{q}) = \bm{J}\bm{J}^\trsp.
	\label{eq:M_base}
\end{equation}

The eigenvector with the maximum eigenvalue of the matrix represents the direction with the greatest velocity that the robot can generate based on its current joint angle configuration and in the limit of joint velocity contained in the unit sphere. A force manipulability can be defined similarly by replacing velocity commands with torque commands.

The above definition of manipulability ignores the physical property of the robot, which can consist of constraints on the commands, or in the form of preferences to set the contribution of each actuator in the kinematic chain. In \cite{jaquier2021geometry}, a diagonal matrix $\bm{W}=\diag (\dot{q}_{1, \max}, \cdots, \dot{q}_{n, \max})$ is considered, whose elements correspond to the maximum joint velocities of the robot. Here, we normalize the maximum velocity of each joint to consider this weight using $\bm{W}=\diag (\dot{q}_{1, \max}, \cdots, \dot{q}_{n, \max}) / \max(\dot{q}_{1, \max}, \cdots, \dot{q}_{n, \max})$.
Therefore, \autoref{eq:M_base} becomes
\begin{equation}
	\tilde{\bm{M}}(\bm{q}) = \bm{J}\bm{W}\bm{W}^\trsp\bm{J}^\trsp,
 \label{eq:IJRR_M}
\end{equation}
which can describe the flexibility of the manipulator in task space by considering its joint angle velocity limits, see \cite{jaquier2021geometry} for details.

There are a variety of indicators to measure manipulability, in which the manipulability index is the most commonly used, expressed as
\begin{equation}
	m = \sqrt{\det (\bm{J}\bm{J}^\trsp)},
\end{equation}
which is proportional to the volume of the manipulability ellipsoid. 
The orientation of the main axis of the ellipsoid depends on the joint angle configuration of the robot. In order to design it according to the required orientation in the task space, transmission ratios \cite{chiu1988task} has first been proposed to measure directional manipulability as a manipulability measure along a specific direction by projecting the ellipsoid onto the direction vector, defined as
\begin{equation}
	\alpha = \sqrt{\bm{u}^\trsp(\bm{J}\bm{J}^\trsp)\bm{u}},
\end{equation}
where $\bm{u}$ represents the direction of interest. $\alpha$ corresponds to the distance along the vector $\bm{u}$ from the origin to the surface of the ellipsoid.

\subsection{Impact aware analysis of the hammering task}
Hammering a nail is a typical example of impact-aware tasks, corresponding to a momentum or energy transfer problem. The difference between humans hammering a nail with and without a tool lies in the dynamic model of the task. Without a tool, the momentum transfer occurs between the human and the nail, while when using a tool, it occurs between the tool and the nail. The dynamic model with the tool does not include the human because of the soft connection between the human and the tool. The role of the human is here to stabilize the tool while absorbing the vibration generated at the moment of impact. 

\begin{figure}
    \centering
    \includegraphics[width=0.6\columnwidth]{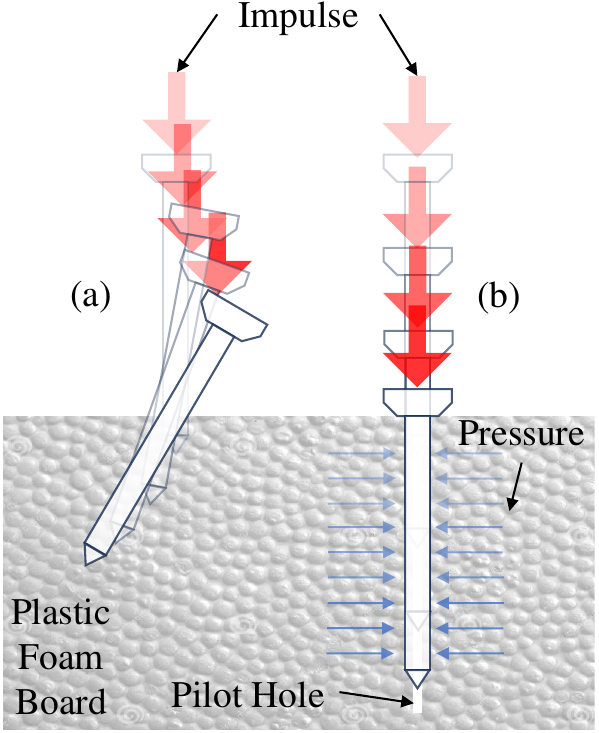}
    \caption{Illustration of a nail driven into a plastic foam. (a) shows the situation without a preformed insertion shape (pilot hole), where sideslip problems can occur; (b) shows the typical situation with a pilot hole, where the nail can be driven more easily so that it remains straight.}
    \label{Fig19}
\end{figure}

We selected a plastic foam material for the hammering task. The material is tight and soft, making it easy to break through. However, the interior of the foam is composed of spherical particles, making the internal material distribution non-uniform, resulting in possible sideslip deviation during the nailing process, as shown in \autoref{Fig19}(a). When considering materials such as wood, concrete, or plastic, pilot holes (smaller than the nails) can be used as a way to drive screws (e.g., for furniture assembly). Despite the presence of a pilot hole, the resistance gradually increases during the insertion process due to the compactness and elasticity of the material, as shown in \autoref{Fig19}(b), where the blue arrow represents the pressure.

According to the above analysis, the impulse model of hammering with a tool is constructed between the hammer and a nail. With the impulse-momentum theorem \cite{halliday2013fundamentals}, the impulse during the hammering task is defined as
\begin{equation}
	\int_{0}^{t}\bm{F}(t)dt = m\bm{v}^h(t) - m\bm{v}^h(0),
\end{equation}
where $\bm{F}(t)$ represents the varying impulsive force along the nail orientation with the duration of the collision $t$. $\bm{v}^h(0)$ and $\bm{v}^h(t)$ represent the velocity along the driving direction of the hammer head before and after the collision.
The impulse-momentum theorem states that the impulse applied to an object will equate to the change in its momentum. With the help of the pilot hole, we can neglect other additional momentum losses such as rotational kinetic energy, plastic deformation, and heat. Even though these losses are still present, the presence of the pilot hole greatly reduces their effects. In addition, as the depth of the nail into the board gradually increases, the force due to the squeezed foam increases, which creates resistance during insertion. Based on the above assumption, in the nail hammering task, we assume $\bm{v}^h(t)=0$. Therefore, for a given hammer mass, the velocity before the collision is the key factor to generate a powerful hammering motion.

\section{Problem formulation}
\label{sec:Definitions}
\subsection{ADMM for Constrained iLQR}
This section uses ADMM optimization to solve an OCP by considering inequality constraints with non-linear dynamical systems and non-quadratic cost functions through an iLQR method. We separated the problem into two parts: 

1) non-convex problem without inequality constraints;

2) an analytical projection step resulting into a fast solver.

The constrained problem is defined as
\begin{equation}
	\begin{array}{rl}
		\underset{\bm{x}_t, \bm{u}_t}{\min} & \sum_{t=0}^{T} c_t(\bm{x}_t) + {\Vert \bm{u}_t \Vert}^2_{\bm{R}_t} \\
		\text{s.t.} & \bm{x}_{t+1} = \bm{f}(\bm{x}_t, \bm{u}_t), \\
		& \mathcal{C}_x:\quad \bm{a}_{x} \leq \bm{l}(\bm{x}_t) \leq \bm{b}_{x}, \\
		& \mathcal{C}_u:\quad \bm{a}_{u} \leq \bm{h}(\bm{u}_t) \leq \bm{b}_{u}, \\
	\end{array}
	\label{eq:ilqr_cost}
\end{equation}
where $c(\bm{x})$ is a non-quadratic function of the state and represents the general formulation of the cost related to the hammering task. We introduce the components of the state cost function in Section \ref{sec:other_cost} (see complete cost function in \autoref{eq:ilqr_cost_final}).
$c(\bm{x},\bm{u})=\sum_{t=0}^{T} c_t(\bm{x}_t) + {\Vert \bm{u}_t \Vert}^2_{\bm{R}_t}$
corresponds to the term $c(\bm{x})$ in \autoref{eq:ADMM_generic}, where the variable $\bm{x}$ in $c(\bm{x})$ covers the system state $\bm{x}$ and control $\bm{u}$ in our problem. 
$\bm{f}(\cdot)$ is a nonlinear function on the state and control variables. We consider the tool affordance as a convex constraint in the task space. For example, the tool affordance of a hammer can be approximated to the prismatic range of the handle that can be grasped in the task space. The state $\bm{x}_t$ is composed of the joint position of manipulator $\bm{q}_t$ and its end-effector position $\bm{p}_t=\bm{f}^\kin(\bm{q}_t)$, where $\bm{f}^\kin(\cdot)$ represents the forward kinematics of the manipulator. Here, we use velocity control, so the control command $\bm{u}_t$ consists of the joint angle velocity $\dot{\bm{q}}_t$. Thus, with this combination of state and control variables, the system is nonlinear. We also added control boundaries for joint velocity limits according to the robot used in the experiment. $\bm{l}(\cdot)$ and $\bm{h}(\cdot)$ represent the general form of the function between the system state and the task state where the constraint occurs, in our case $\bm{l}(\bm{x}_t)=\bm{p}_t$ and $\bm{h}(\bm{u}_t)=\bm{u}_t$. We define $\mathcal{C}_x$ and $\mathcal{C}_u$ as the inequality constraint on the state and control variables, respectively.

\subsubsection{Step 1 of ADMM}
To solve the above optimization, we can separate the inequality constraints from the problem according to \autoref{eq:admm_convex} and solve a regularized version of the problem in \autoref{eq:ilqr_cost} without inequality constraints as
\begin{equation}
	\begin{array}{rl}
		\underset{\bm{x}_t, \bm{u}_t}{\min} & \sum_{t=0}^{T} c_t(\bm{x}_t) + {\Vert \bm{u}_t \Vert}^2_{\bm{R}_t} + \frac{\rho_x}{2}{\Vert \bm{x}_t-\bm{z}_{x,t} + \bm{\lambda}_{x,t}\Vert}^2\\
		& \hspace{34mm} + \frac{\rho_u}{2}{\Vert \bm{u}_t-\bm{z}_{u,t}+\bm{\lambda}_{u,t} \Vert}^2\\
		\text{s.t.} & \bm{x}_{t+1} = \bm{f}(\bm{x}_t, \bm{u}_t),
	\end{array}
	\label{eq:ilqr_cost_constraint_consensus}
\end{equation}
where the update of $\bm{z}_{x/u}$ is the Euclidean projection onto the state and control boundary and $\bm{\lambda}_{x/u}$ is the scaled dual variable as in \autoref{eq:admm_scaled}.

The constraints for the state of the end-effector position can be defined to occur at any timestep, which in this article corresponds to the timestep of picking up the tool. Therefore, the projection on the state inequality constraint $\mathcal{C}_x$ is active when the pick-up action occurs. In addition, the projection of the control inequality constraint $\mathcal{C}_u$ is valid throughout the motion. We simplify the above formula by adding the following abbreviation to provide compact expressions. $\bm{Q}_{r,t}$ and $\bm{R}_{r,t}$ are penalty parameters in the form of diagonal matrices for the state $\bm{x}_t$ and control command $\bm{u}_t$.
We define $\bm{x}_{r,t}=\bm{z}_{x,t}-\bm{\lambda}_{x,t}, \bm{u}_{r,t}=\bm{z}_{u,t}-\bm{\lambda}_{u,t}.$ Then, we get the following expression
\begin{equation}
	\begin{array}{rl}
		\underset{\bm{x}_t, \bm{u}_t}{\min} & \sum_{t=0}^{T} c_t(\bm{x}_t) + {\Vert \bm{u}_t \Vert}^2_{\bm{R}_t} +{\Vert \bm{x}_t-\bm{x}_{r,t} \Vert}^2_{\bm{Q}_{r,t}}  \\
		& \hspace{34mm} + {\Vert \bm{u}_t-\bm{u}_{r,t} \Vert}^2_{\bm{R}_{r,t}}\\
		\text{s.t.} & \bm{x}_{t+1} = \bm{f}(\bm{x}_t, \bm{u}_t).
	\end{array}
	\label{eq:ilqr_cost_constraint}
\end{equation}

We perform a first-order Taylor expansion of the dynamical system $\bm{x}_{t+1}=\bm{f}(\bm{x}_t, \bm{u}_t)$ around some nominal realization of the plant denoted as $(\bm{\hat{x}}_t, \bm{\hat{u}}_t)$, namely
\begin{align}
		\bm{x}_{t+1} &\approx \bm{f}(\bm{\hat{x}}_t, \bm{\hat{u}}_t) +  \frac {\partial \bm{f}}{\partial \bm{x}_t} (\bm{x}_t - \bm{\hat{x}}_t)+ \frac {\partial \bm{f}}{\partial \bm{u}_t} (\bm{u}_t - \bm{\hat{u}}_t) \nonumber\\
		\iff \Delta \bm{x}_{t+1} &\approx \bm{A}_t \Delta \bm{x}_{t} + \bm{B}_t \Delta \bm{u}_{t},
\end{align}
with $\Delta \bm{x}_{t} = \bm{x}_t - \bm{\hat{x}}_t, \Delta \bm{u}_{t} = \bm{u}_t - \bm{\hat{u}}_t$, and Jacobian matrices $\bm{A}_t=\frac {\partial \bm{f}}{\partial \bm{x}_t}\big|_{\bm{\hat{x}}_t, \bm{\hat{u}}_t}$, $\bm{B}_t=\frac {\partial \bm{f}}{\partial \bm{u}_t}\big|_{\bm{\hat{x}}_t, \bm{\hat{u}}_t}$. The concatenated form of the linearized dynamics model can be written as $\Delta \bm{x}=\bm{S}_{\bm{x}} \Delta\bm{x}_1+\bm{S}_{\bm{u}} \Delta \bm{u}$. Since in our problem, we start from $\Delta\bm{x}_1 = 0$, therefore, $\Delta \bm{x}=\bm{S}_{\bm{u}} \Delta \bm{u}$, see  Appendix for details.
We then approximate $c(\bm{x}_t)$ by a second-order Taylor expansion around $\bm{\hat{x}}_t$, namely
\begin{equation}
	\begin{aligned}
		c_t\left(\bm{x}_t\right) & \approx c_t\left(\bm{\hat{x}}_t\right)+{\bm{k}^{\bm{x}}_t}^\trsp\left(\bm{x}_t-\bm{\hat{x}}_t\right)\\
		&\hspace{18mm}+\frac{1}{2}\left(\bm{x}_t-\bm{\hat{x}}_t\right)^{\trsp}\bm{K}^{\bm{x}\bm{x}}_t\left(\bm{x}_t-\hat{\bm{x}}_t\right) \\
		& \approx \frac{1}{2}\left(\Delta \bm{x}_t-\bm{\mu}_t\right)^{\trsp} \bm{K}^{\bm{x}\bm{x}}_t\left(\Delta \bm{x}_t-\bm{\mu}_t\right)+\text{const.},
	\end{aligned}
\end{equation}
where $\bm{k}^{\bm{x}}_t = \nabla _{\bm{x}_t} c_t(\bm{x}_t)$ and $\bm{K}^{\bm{x}\bm{x}}_t=\nabla ^2 _{\bm{x}_t} c_t(\bm{x}_t)$ are the Jacobian and the Hessian matrices, and $\bm{\mu}_t=-{\bm{K}^{\bm{x}\bm{x}}_t}^{-1} \bm{k}^{\bm{x}}_t$. 

With the above approximation, the cost of the state $c(\bm{x})$ can be changed from a non-convex to a convex form.
Then, the cost $c(\bm{x})$ embedding the control part can be rewritten in batch form by removing the constant terms as $c(\bm{x}, \bm{u}) = \frac{1}{2}{\Vert \Delta\bm{x}-\bm{x}_d \Vert}^2_{\bm{K}^{\bm{x}\bm{x}}_t} + {\Vert \Delta\bm{u}-\bm{u}_d \Vert}^2_{\bm{R}}$, with $\bm{u}_d=-\bm{\hat{u}}$. We define $\Delta\bm{x}_r=-(\hat{\bm{x}}-\bm{x}_r)$, $\Delta\bm{u}_r=-(\hat{\bm{u}}-\bm{u}_r) $, where variables without indices $t$ denote the concatenation of the variables for the whole task duration. One iteration of iLQR consists of the following problem:
\begin{equation}
	\begin{array}{rl}
		\underset{\Delta\bm{x},\Delta \bm{u}}{\min} & \frac{1}{2}{\Vert \Delta\bm{x}-\bm{x}_d \Vert}^2_{\bm{K}^{\bm{x}\bm{x}}} + {\Vert \Delta\bm{u}-\bm{u}_d \Vert}^2_{\bm{R}} \\
		& \qquad + {\Vert \Delta\bm{x}-\Delta\bm{x}_r \Vert}^2_{\bm{Q}_r} + {\Vert \Delta\bm{u}-\Delta\bm{u}_r \Vert}^2_{\bm{R}_r} \\
		\text{s.t.} & \Delta \bm{x} = \bm{S}_{\bm{u}}\Delta \bm{u}.
	\end{array}
	\label{eq:cost_ilqr_admm}
\end{equation}

For such class of problems, the solution can be obtained analytically. This problem follows a similar LQR formulation as \autoref{eq:batch_cost} in the Appendix. We focus here on the batch least-squares solution of LQR. Its analytic solution is given by 
\begin{equation}
	\begin{array}{rl}
		\Delta\bm{\hat{u}} &= \left(\bm{S}_{\bm{u}}^\trsp\left(\frac{1}{2}\bm{K}^{\bm{x}\bm{x}}+\bm{Q}_r\right)\bm{S}_{\bm{u}}+\bm{R}+\bm{R}_r\right)^{-1} \\
		&\hspace{6mm} (\frac{1}{2}\bm{S}_{\bm{u}}^\trsp\bm{K}^{\bm{x}\bm{x}}\bm{x}_d+\bm{S}_{\bm{u}}^\trsp\bm{Q}_r\Delta \bm{x}_r+\bm{R}_r\Delta \bm{u}_r).
	\end{array}
	\label{eq:analytical solution}
\end{equation}

Although the problem in this paper is solved using an approximation into a convex problem, the optimization problem for the original problem is still non-convex. Therefore, with the help of the ADMM algorithm in solving such non-convex problems, it does not need to give a feasible solution.

Denoting the solution to \autoref{eq:cost_ilqr_admm} at iteration $k_j$ as $\Delta \bm{\hat{u}}_{k_j}$ and given the current nominal state and control $\left\{\bm{\hat{x}}^{k_i}_{k_j}, \bm{\hat{u}}^{k_i}_{k_j}\right\}$, we perform a line search as  in Algorithm \ref{alg:linesearch} to determine the next nominal state and control command $\left\{\bm{\hat{x}}^{k_i}_{k_j+1}, \bm{\hat{u}}^{k_i}_{k_j+1}\right\}$ in the iLQR loop, where $\bm{x}=F(\bm{u})$ is the forward pass function in vector form, built from $\bm{x}_{t+1} = \bm{f}(\bm{x}_t, \bm{u}_t)$. $k_i$ and $k_j$ represent the iteration indices in the ADMM and iLQR loops, respectively (see Appendix and \cite{lembono2021probabilistic} for details of the derivation).

\subsubsection{Step 2 of ADMM}
After Step 1 of ADMM for iLQR, we obtain the converged nominal solution $\left\{\bm{\hat{x}}^{k_i+1}, \bm{\hat{u}}^{k_i+1}\right\}$, where the projection step is derived below.
In the case of control bounds, $\bm{h}(\bm{u})= \bm{u}$ and the projection step corresponds to clipping the values of $\bm{u}$ between the bounds $\bm{a}_u$ and $\bm{b}_u$, as shown in the Appendix. The projection is $\Pi_{\mathcal{C}_u}(\bm{u})=\text{clip}(\bm{u}, \text{lower}=\bm{a}_u, \text{upper}=\bm{b}_u)$. In the case of state bounds, $\bm{l}(\bm{x})= \bm{x}$ and the projection step corresponds to clipping the values of $\bm{x}$ between the bounds $\bm{a}_x$ and $\bm{b}_x$, as shown in the Appendix. The projection is $\Pi_{\mathcal{C}_x}(\bm{x})=\text{clip}(\bm{x}, \text{lower}=\bm{a}_x, \text{upper}=\bm{b}_x)$. Above all, the projection step ensures that the solution remains feasible with respect to the hard constraints. If a candidate solution violates the constraints, the projection step clips it to satisfy the constraints. Then, the hard constraints problem can be naturally handled in the dual update step. The state projections in our work correspond to the tool affordance range represented by a prismatic constraint. The whole ADMM-iLQR algorithm process is summarized in Algorithm \ref{alg:1}.

\begin{algorithm}[tb]
	\setstretch{1.1}
	\caption{ADMM with constrained iLQR}\label{alg:1}
	\KwData{Set $\bm{Q}_r, \bm{R}_r, k_{\text{max}}$, primal and dual residual thresholds $r_{p, \text{max}}, r_{d, \text{max}}, c_{\text{max}}$; \\
		\quad \qquad Initialize $k=0, \bm{z}^0_x=\bm{0}, \bm{\lambda}^0_x=\bm{0}, \bm{z}^0_u=\bm{0}, \bm{\lambda}^0_u=\bm{0}$ and $r^0_p \geq r_{p, \text{max}}, r^0_d \geq r_{d, \text{max}}$ \\
		\quad \qquad Initialize the nominal state $\bm{\hat{x}}^0$ and control $\bm{\hat{u}}^0$}
	\KwResult{optimal $\bm{x}^*, \bm{u}^*$}
	\While{$k_i < k_{\text{max}}^{\text{ADMM}}$ and  $r^k_p > r_{p, \text{max}}$ and $r^k_d > r_{d, \text{max}}$}{
        \While{$k_j < k_{\text{max}}^{\text{iLQR}}$ and $c(\hat{\bm{x}}_{k_j}^{k_i}, \hat{\bm{u}}_{k_j}^{k_i})>c_{\text{max}}$}{Solve \autoref{eq:cost_ilqr_admm} for $\Delta \bm{u}^{k_i+1}, \Delta \bm{x}^{k_i+1}$ with $\bm{x}_r^{k_i}=\bm{z}^{k_i}_x-\bm{\lambda}^{k_i}_x$ and $\bm{u}_r^{k_i}=\bm{z}^{k_i}_u-\bm{\lambda}^{k_i}_u$;\\
        Do line search to find $\bm{\hat{u}}_{k_j}^{k_i+1}$ and $\bm{\hat{x}}_{k_j}^{k_i+1}$;}
		$\bm{z}^{k_i+1}_x=\Pi_{\mathcal{C}_x}(\bm{\hat{x}}^{k_i+1}+\bm{\lambda}^{k_i}_x)$;\\
		$\bm{z}^{k_i+1}_u=\Pi_{\mathcal{C}_u}(\bm{\hat{u}}^{k_i+1}+\bm{\lambda}^{k_i}_u)$;\\
		$\bm{\lambda}^{k_i+1}_x=\bm{\lambda}^{k_i}_x+\bm{\hat{x}}^{k_i+1}-\bm{z}^{k_i+1}_x$;\\
		$\bm{\lambda}^{k_i+1}_u=\bm{\lambda}^{k_i}_u+\bm{\hat{u}}^{k_i+1}-\bm{z}^{k_i+1}_u$;\\
		$r^{k_i+1}_p={\Vert \bm{\hat{u}}^{k_i+1}-\bm{z}^{k_i+1}_u \Vert}^2_2 + {\Vert \bm{\hat{x}}^{k_i+1}-\bm{z}^{k_i+1}_x \Vert}^2_2$;\\
		$r^{k_i+1}_d={\Vert \bm{z}^{k_i}_x-\bm{z}^{k_i+1}_x \Vert}^2_2 + {\Vert \bm{z}^{k_i}_u-\bm{z}^{k_i+1}_u \Vert}^2_2$.
  }
\end{algorithm}

\begin{algorithm}[tb]
\caption{Line search method with parameter $\alpha_{\min}$}
\label{alg:linesearch}
$\alpha \gets 1$ \\
\While{$c(F(\bm{\hat{u}}+\alpha\;\Delta\bm{\hat{u}}), \bm{\hat{u}}+\alpha\;\Delta\bm{\hat{u}}) > c(\hat{\bm{x}}, \bm{\hat{u}}) \;\textbf{and}\;\; \alpha > \alpha_{\min}$}{
	$\alpha \gets \frac{\alpha}{2}$
}
\end{algorithm}

\subsection{ADMM-iLQR on tool-use motion planning}
We consider the tool-use motion planning problem as a viapoint task with a desired range. We use the batch iLQR form in our experiment by considering joint angle velocity commands, with the dynamics expressed as a single integrator to update the joint angle states given velocity commands. A common and intuitive way to incorporate constraints into an optimization problem is to use soft constraints as a weighted cost. For redundant tasks, the constraints can be satisfied for a large range of weights and do not need fine tuning of the weights. However, for more complex tasks requiring trade-offs, the weights need to be adjusted until a satisfactory result is obtained. In such a case, with this adjustment through weights, it can be difficult to ensure that this constraint is strictly satisfied. The tool affordance in our work can be taken as a constraint in the task space at a specific picking time. Since the tool handling is a necessary prerequisite for the completion of the entire manipulation task, it must be taken as a hard constraint. However, the control commands are described in the joint space of the manipulator. We introduce the tool affordance constraint as a hard constraint into the task space and embed the end-effector position into the state function. Thus, the state function consists of joint position and end-effector position $\bm{x}=\{\bm{q}, \bm{p}\}$. For a robot with $D$ articulations, the system consists of a large vector $\{\bm{q}, \bm{p}\}$ lengths $(D+3)T$ with control commands length $DT$, by concatenating the joint angle states $\bm{q}_t$ and end-effector states $\bm{p}_t$ at different time steps $t\in\{1, \cdots, T\}$.

In \autoref{eq:ilqr_cost}, the state $\bm{l}(\bm{x})$ representing the inequality constraint of the problem is defined as the range of the handle which is represented as a prismatic shape, and the affine projection is used to solve this kind of constraint in the task space to ensure that the robot reaches the hammer handle within the available range. Besides the basic requirement of the position reaching cost, the cost term is also composed of the task requirement, including the orientation of grasping, the final hammering direction and the maximum directional manipulability (introduced in the following subsection).
%

\subsection{Tool usage as an extended link in the robot kinematic chain}
\begin{figure}
	\centering
	\subfigure[Same joint angle configuration with different tool grasps]{
			\centering
			\includegraphics[width=\columnwidth]{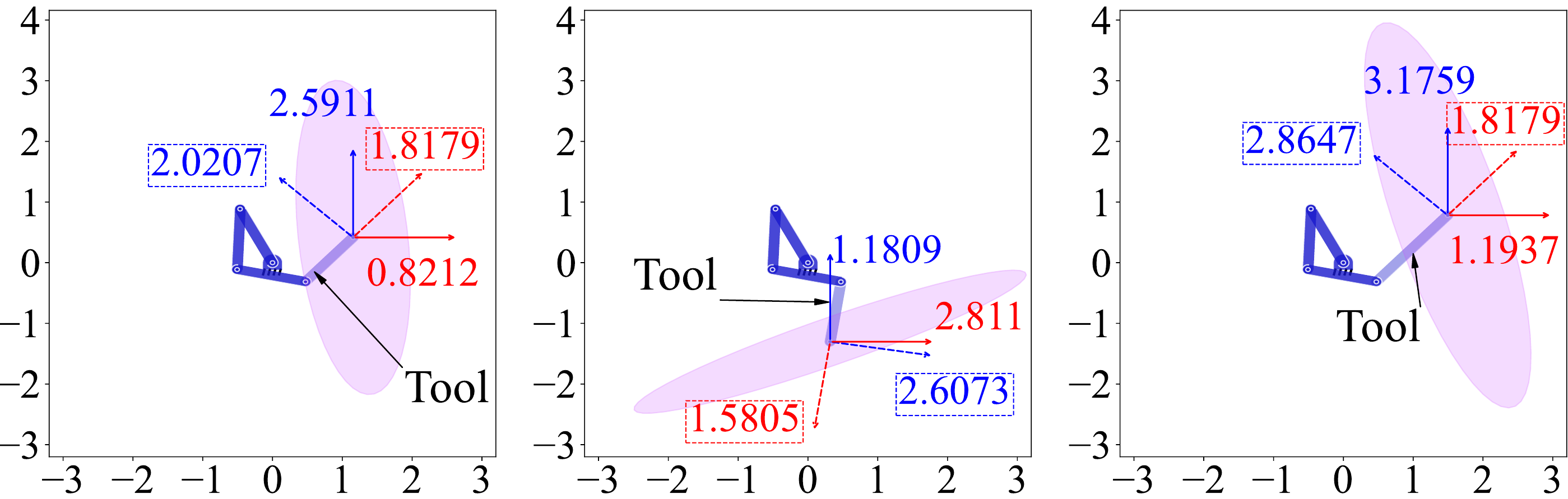}	
			\label{Fig1_1}
		}
		\subfigure[Same tool grasp with different joint angle configurations]{
		\centering
		\includegraphics[width=\columnwidth]{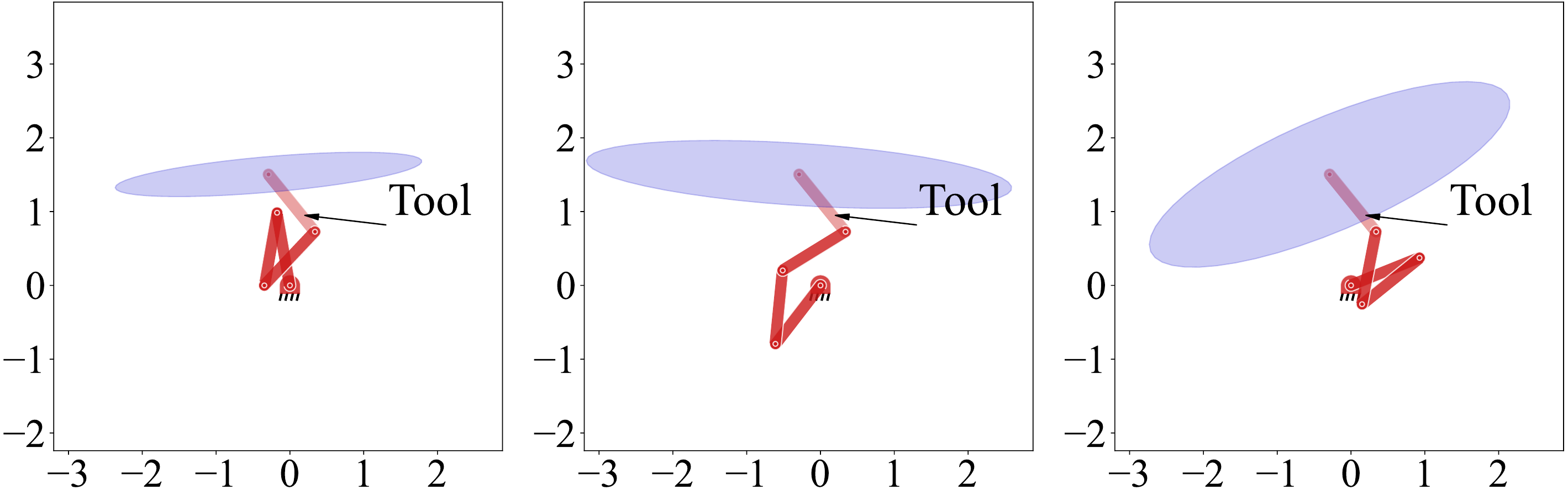}	
		\label{Fig1_2}
	}
		\caption{Influence of the grasping pose on the velocity manipulability of a 3-axis planar arm, where the last link with light color represents the tool. In subfigure (a), a planar arm grasps a tool with the same posture but a different grasping position (left and right plot) and grasping orientation (left and middle plot). The number shows the projection value along the world frame and local end-effector frame (dashed box). Subfigure (b) shows the robot grasping the same tool with different joint angle configurations.}
	\label{Fig1}
	\vspace{-0.3cm}
\end{figure}

After a tool is grasped, the tool can be seen as an external link to the robot, where the new end-effector is closely related to the notion of tool affordance. Before the hammering motion is executed, the soft connection can be seen as a fixed joint, in which the original kinematic chain changes depending on the grasping location. When handling tools, the manipulability of the end-effector of the robot (e.g., gripper) and the end-effector of the tool it holds can change considerably. This kinematic chain can be brought into the optimal control problem to maximize the manipulability of the extended kinematic chain. As shown in \autoref{Fig1}, the gripping point location along the tool and the gripper pose used to grasp this tool can result in drastically different manipulability. In contrast, if we use the same grasping orientation, the projection along the end-effector direction is constant, independently of the grasping point on the tool. This phenomenon can be observed by the projection value in \autoref{Fig1_1}.

\subsection{Maximization of Velocity Manipulability}
In this section, we discuss the different manipulability metrics on the directional task.
Directional manipulability is an appropriate metric for the measurement and optimization of manipulability in a particular direction. In addition, there are also other commonly used metrics for measuring the manipulability of manipulators.
The method for maximizing the end-effector volume of a manipulability ellipsoid is the most commonly used metric, which does not take into account tool affordance and directionality. Its isotropic expansion can typically lead to failure to meet the specified direction required by the task. In the following, we abbreviate this approach as \emph{Common Metric}. The third method is to set a desired ellipsoid whose main axis is large in the desired direction, as in \cite{jaquier2021geometry}. This approach cannot guarantee that the manually specified ellipsoid is reachable by the robot. The robot will also consider that matching the secondary axes is equally important as matching the main axis of the ellipsoid.
We introduce the above three methods in the comparative study, with respective cost functions defined as
\begin{align}
	c^1_{\text{man},T} &= w_\text{man}\norm{\bm{W}^\trsp\bm{J}(\bm{q}_T)^\trsp\bm{n}}^{-2}_2,\\
	c^2_{\text{man},T} &= w_\text{man}\left(\det (\bm{J}(\bm{q}_T)\bm{W}\bm{W}^\trsp\bm{J}(\bm{q}_T)^\trsp)\right)^{-2},\\
	c^3_{\text{man},T} &= w_\text{man}\norm{\log(\bm{M}_d^{-\frac{1}{2}}\tilde{\bm{M}}(\bm{q}_T)\bm{M}_d^{-\frac{1}{2}})}^2_\text{F},
\end{align}
where $c_{\text{man}, T}^{\{1, 2, 3\}}$ represents the cost of different strategies for maximizing manipulability (respectively, directional manipulability, \emph{common metric} and tracking of desired manipulability) at the final time step $T$, and at other time steps $c_{\text{man}, T}=0$. $w_\text{man}$ represents the weight in the cost function.
$\bm{n}\in\mathbb{R}^3$ in {$c_{\text{man}, T}^1$} represents the desired direction of maximization in the task space. $\bm{W}$ represents the velocity capacity of each joint and $\tilde{\bm{M}}(\bm{q}_T)$ represents the manipulability at the final time step $T$. Both values are defined in \autoref{eq:IJRR_M}. $\bm{M}_d$ represents the desired manipulability (set manually with an axis having a large value along the task direction).

\subsection{Other task requirements defined in the cost function}
\label{sec:other_cost}
We simplify the optimization problem by treating orientation constraints as soft constraints in the cost function by assigning appropriate weights to them.
For a two-finger gripper used as robot hand, the best way to grasp a cylindrical object is to keep the grasping direction perpendicular to its axis, ensuring the hammer does not fall off during transportation. Therefore, to define a suitable viapoint with a desired grasping range, we add an orientation constraint 
\begin{equation}
    c_{\text{o},t'} = w_\text{o} e_\text{o}^2, \quad\text{with}\quad 
    e_\text{o} = {\bm{R}^y_r}^\trsp \bm{R}^z_h,\quad
\end{equation}
where $\bm{R}$ with $y$ and $z$ superscripts represent $y$ and $z$ column vector of the rotation matrix, the subscript $r$ represents robot and $h$ represents the hammer. $c_{\text{o},t'}$ is used to force the gripper grasping direction $y$ to be perpendicular to the axis $z$ of the hammer handle at hammer picking time step $t'$, and for other time step $c_{\text{o},t}=0$. $w_\text{o}$ is the weight of the cost. 

Before hammering, the robot should take the hammer to the accurate desired pose at the final time step $T$, where the hammer should be above the nail with the direction of the head parallel to the nail. For the position part, the corresponding cost with weight $w_\text{pos}$ is defined as 
\begin{equation}
    c_{\text{pos},T} = w_\text{pos} \bm{e}_\text{pos}^\trsp \bm{e}_\text{pos},
    \quad\text{with}\quad 
    \bm{e}_\text{pos} = \bm{f}(\bm{q}_T)-\bm{f}_d.
\end{equation}

The direction vector, which is a unit vector in three-dimensional space, can be viewed as a point in a Riemannian manifold $\mathcal{S}^2$, representing a unit sphere. The distance between two directions is computed on $\mathcal{S}^2$, see Appendix for details. The corresponding cost with weight $w_\text{dir}$ is
\begin{equation}
    \begin{aligned}
    c_{\text{dir},T} &= w_\text{dir} \bm{e}^\trsp_\text{dir} \bm{e}_\text{dir}, \quad\text{with}\\
		\bm{e}_\text{dir} &= \text{Log}^{\mathcal{S}^2}_{\bm{v}^h}(\bm{v}^r_T) = \arccos({\bm{v}^h}^\trsp \bm{v}^r_T) \, \frac{\bm{v}^r_T- {\bm{v}^h}^\trsp \bm{v}^r_T\, \bm{v}^h}{\|\bm{v}^r_T - {\bm{v}^h}^\trsp \bm{v}^r_T \, \bm{v}^h\|}.
  \end{aligned}
\end{equation}

\begin{figure*}
	\centering
	\subfigure[2D space with 3-DoF planar robot]{	
		\centering
		\includegraphics[width=0.7\linewidth]{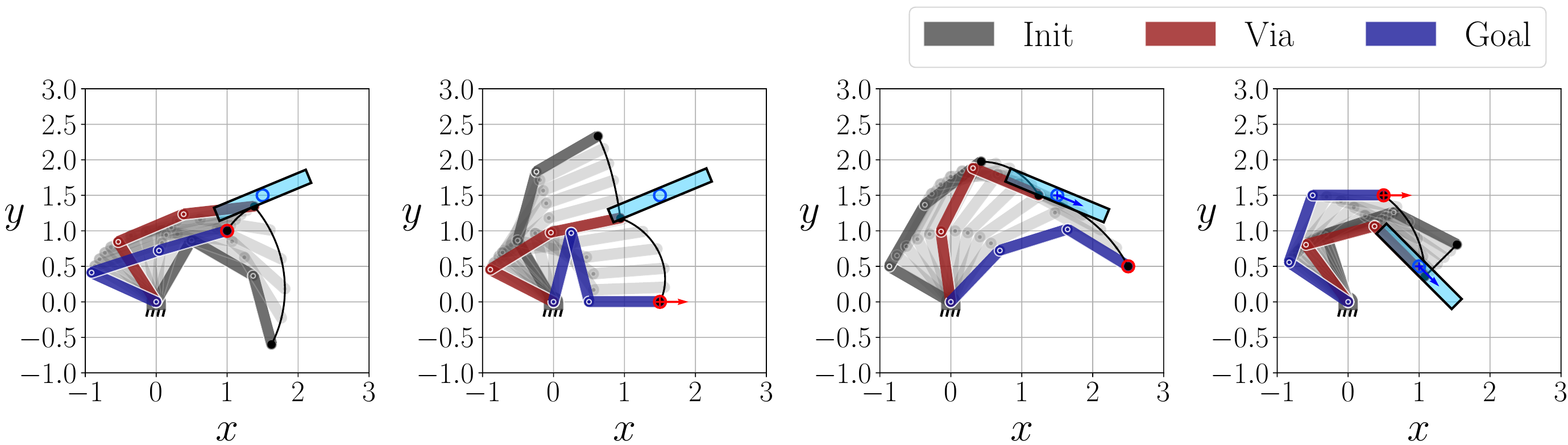}	
		\label{Fig3_1}
	}
	\subfigure[3D space with Franka Emika 7-DoF robot]{	
	\centering
	\includegraphics[width=0.25\linewidth]{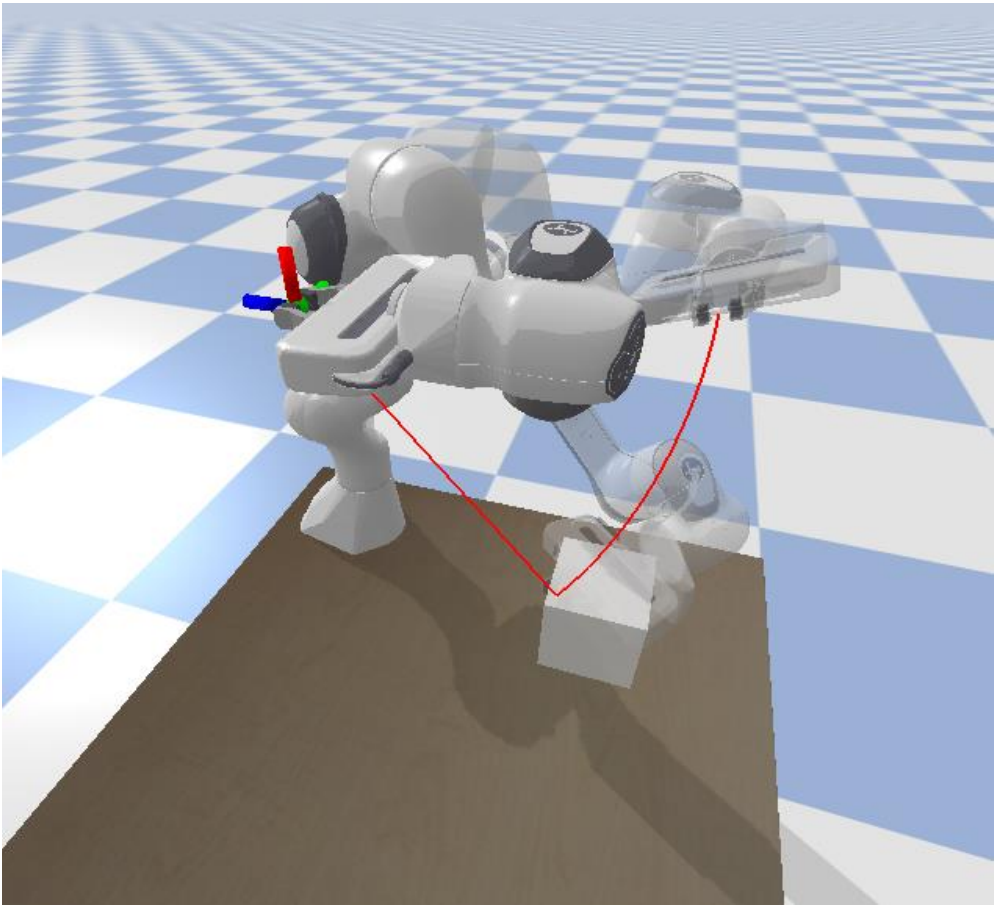}	
	\label{Fig3_2}
}
	\caption{ADMM-iLQR applied to a viapoint task with a desired range task. In subplot (a), we show four examples of the viapoint with a desired range, where the grey, red and blue robots represent the initial, via and final states of the robot, respectively. The first plot requires the robot to reach the range in cyan and then the final position (red ring). We increase the complexity of the task by setting the final pose specified after passing through the range, arriving at the range while maintaining a desired pose, and finally combining both constraints above into the fourth task scenario. The arrows in the last three plots (blue: viapoint with a desired range; red: final) represent the specified direction. In subplot (b), our approach is verified on a simulated Franka Emika robot.}
	\label{Fig3}
	\vspace{-0.3cm}
\end{figure*}

We also need to consider the physical joint limits during motion planning. These joint angle constraints are usually viewed as a hard constraint in optimization problems. However, it can also restrict the exploration of joint configuration possibilities during iteration, especially when exploring maximum manipulability, where joint angle configurations significantly impact the final result. We noticed that taking joint physical limits as hard constraints can easily make the solution fall into a local optimum. Soft constraints have thus been considered for this part of the cost, represented by an activation function as weight in the cost function, shown as 
\begin{equation}
	\begin{array}{rl}
		c_{\text{lim},t} &= \norm{\bm{q}_L-\bm{q_t}}^2_{\bm{\Lambda}_t}, \quad\text{with}\\
		\Lambda_t^i &=  \left\{
		\begin{aligned}
			1 & \quad \text{if} \quad  \bm{q}_t^i \geq \max (\bm{q}_{L_i}) \quad \text{or} \quad \bm{q}_t^i \leq \min (\bm{q}_{L_i}) \\
			0 & \quad \text{otherwise}
		\end{aligned}
		\right.
	\end{array}
\end{equation}
so that the constraint cost is activated if the joint position exceeds the limit, where this cost $c_\text{lim}$ is valid throughout the timeline.

Summarizing the presentation of all the above constraints and cost functions, the final optimization problem related to \autoref{eq:ilqr_cost} becomes
\begin{equation}
	\begin{array}{rl}
		\underset{\bm{x}_t, \bm{u}_t}{\min} & \sum_{t=0}^{T} c_{\text{o},t}(\bm{x}_t) + c_{\text{pos},t}(\bm{x}_t) + c_{\text{dir},t}(\bm{x}_t) \\
        & \hspace{10mm} + c_{\text{lim},t}(\bm{x}_t) + c_{\text{man},t}(\bm{x}_t) + \norm{\bm{u}_t}^2_{\bm{R}} \\
		\text{s.t.} & \bm{x}_{t+1} = \bm{f}(\bm{x}_t, \bm{u}_t), \\
		& {\mathcal{C}_x:} \bm{a}_{x} \leq \bm{l}(\bm{x}_t) \leq \bm{b}_{x}, \\
		& {\mathcal{C}_u:} \bm{a}_{u} \leq \bm{h}(\bm{u}_t) \leq \bm{b}_{u}, \\
	\end{array}
	\label{eq:ilqr_cost_final}
\end{equation}
where $\{\bm{a}_x, \bm{b}_x\}$ represents the prismatic constraint of the tool affordance range occurring at the tool pick-up timestep, and $\{\bm{a}_u, \bm{b}_u\}$ represents the joint velocity limit of the manipulator through the entire motion.

\section{Simulation}
\label{sec:simulation}
To validate ADMM-iLQR in the task space constrained motion planning, we design a viapoint task with a desired range and extend its application to the pick-and-place task in the 2D and 3D space on a planar robot and Franka Emika robot in simulation. Then, we introduce the maximization of manipulability along a desired direction to the optimal control problem and show the difference in manipulability metrics. The optimization algorithm is implemented in Python 3.8, running on a personal computer with Intel(R) i7-10750H CPU 2.60GHz.

\begin{figure*}[htbp]
	\centering
	\subfigure[Pick-and-place task with a 3-axis planar robot]{	
		\centering
		\includegraphics[width=0.65\linewidth]{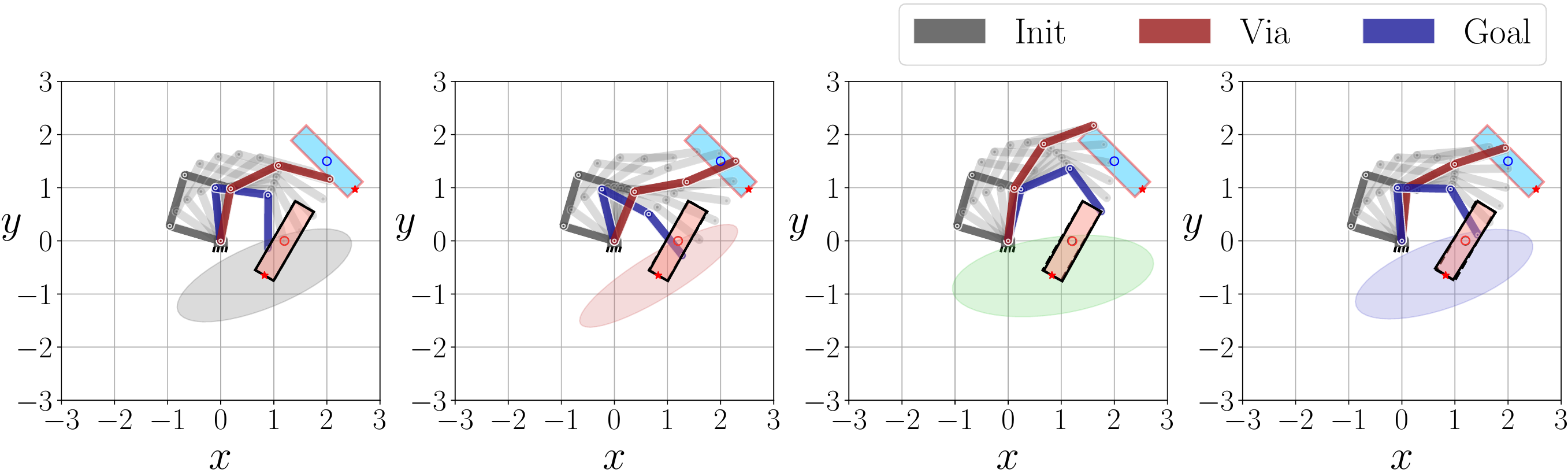}	
		\label{Fig4_1}
	}
	\subfigure[Velocity manipulability at the tip]{	
		\centering
		\includegraphics[width=0.3\linewidth]{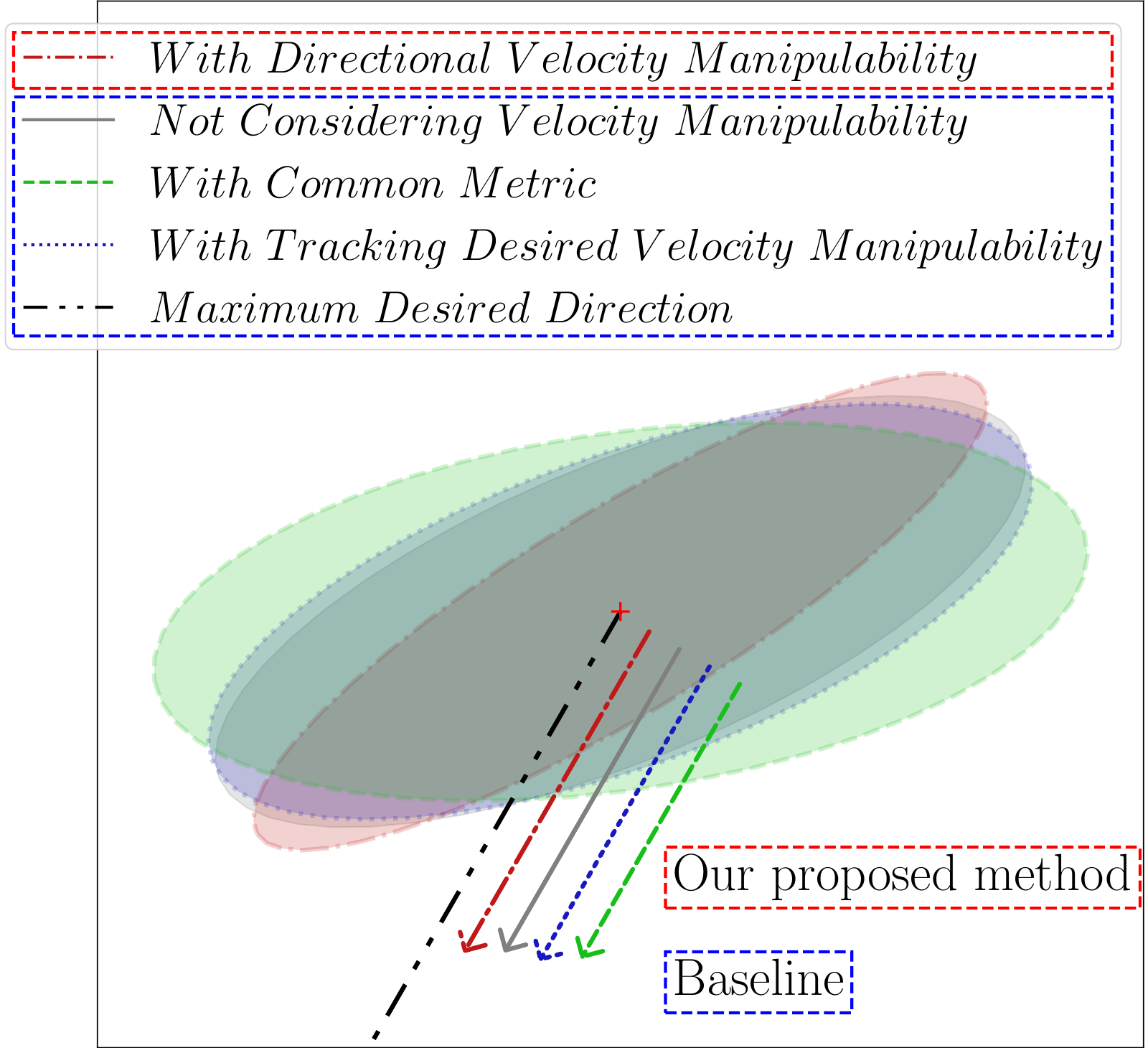}	
		\label{Fig4_2}
	}
	\caption{ADMM-iLQR applied to a pick-and-place task. In (a), we show the results of considering the maximization of manipulability, where the grey, red and blue robots correspond to the descriptions in \autoref{Fig3_1}. The robot is required to pick the blue cube and place it in the black dashed cube region while maximizing the velocity manipulability of the robot-tool system along the object orientation. The arrow in the last plot represents the desired direction and the red dot represents the object tip. In (b), we display the velocity manipulability of the robot-tool system separately on the tool tip generated by different strategies. The projections along the desired direction generated by the different approaches are represented with arrows of different line styles. To observe the differences between each projection more clearly, we lined them up in parallel. For a better view of the overlayed ellipses, see the color version of this article.}
	\label{Fig4}
	\vspace{-0.3cm}
\end{figure*}

\subsection{ADMM-iLQR in a viapoint task with a desired range}
\label{sec:simulation_admm_ilqr_via_range}
We define a square range at the intermediate timestep ($t'=T/2$) to simulate the available range to grasp the hammer handle, then set a desired pose at the final timestep, as shown in \autoref{Fig3}. In the planar task, the length of each link of the planar robot is $1.0$, and the size of the available range for the hammer handle is defined as a rectangular shape with length $1.4$ and width $0.2$. The control boundary is defined as $[-4, 4]$(rad/s) for the 2D examples and as $[-3, 3]$(rad/s) for the 3D examples. In the 3D simulation, we import the URDF model of the Frank Emika robot, where the available range is defined as a cube with size $1$. 
The initial control states in both cases are set as $\bm{u}=\bm{0}$. In the iLQR parameter setting, the final pose reaching weight is \{\emph{2D}-$\text{position and orientation: }1e2$; \emph{3D}-$\text{position: }1e1,\text{orientation: }1e0$\}, and the control weight for both spaces is $\bm{R}=\diag(1e\textnormal{-}5)$. The cost threshold is $c_{max}=1$. The penalty parameter for state and control of the ADMM are set as \{$\bm{Q}_r$-\emph{2D}: $\diag(1e1)$; \emph{3D}: $\diag(1e0)$\}, \{$\bm{R}_r$-\emph{2D}/\emph{3D}: $\diag (1e\textnormal{-}3)\}$, respectively. The threshold is $r_{p, max}=r_{d,max}=1e\textnormal{-}4$. Also, the time step is \{$dt$-\emph{2D}: $0.01$s; \emph{3D}: $0.06$s\}, and the trajectory horizon is $T=100$. The maximum iteration numbers of the iLQR algorithm and ADMM algorithm are $k_{max}^{iLQR}=10$ and $k_{max}^{ADMM}=20$, respectively. We initialize the robot close to this range. In the first task, we can observe that the robot can reach the range and the target position. We then add the orientation constraints to increase the complexity of the task, where the robot needs to maintain a specified orientation at the viapoint (with a desired range) and at the final state. The results show that the ADMM-iLQR approach performs well in solving OCP with the state constraint in task space. The computation time in the four scenarios of \autoref{Fig3_1} is $10.03$s, $21.43$s, $18.57$s and $5.34$s, respectively. The planar robot can find an optimal joint angle trajectory depending on the relations between its initial configuration, the viapoint with a desired range and the final target, effectively anticipating the grasp by exploiting tool affordance. 

\subsection{ADMM-iLQR with maximization of manipulability in a pick-and-place task}
\label{sec:pick_place_mani}
\begin{figure}[htbp]
    \centering
    \includegraphics[width=\columnwidth]{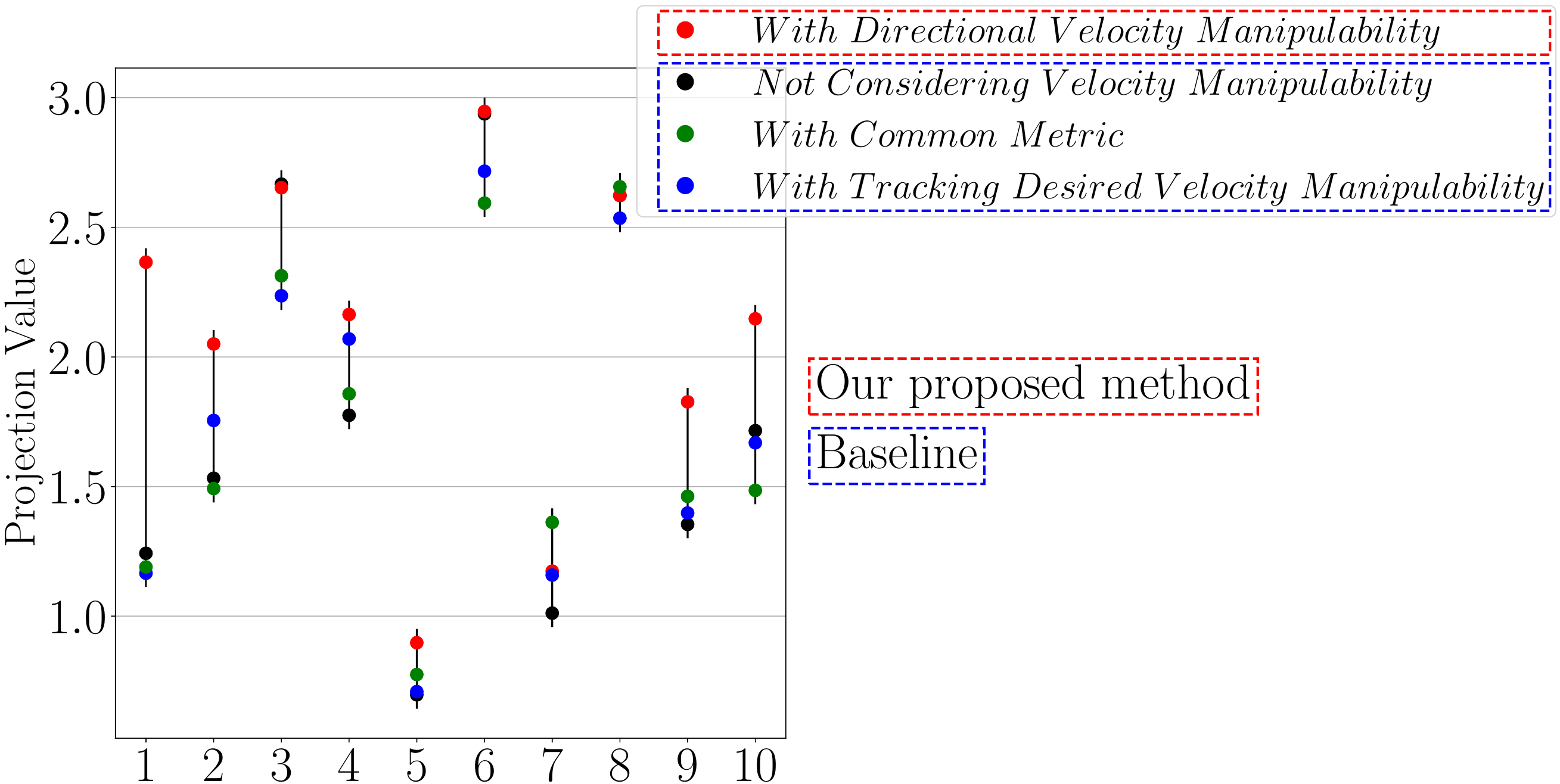}
    \caption{Projection of the velocity manipulability along the desired direction in ten simulations.}
    \label{Fig12}
\end{figure}

\begin{figure}[htbp]
    \centering
    \includegraphics[width=0.9\columnwidth]{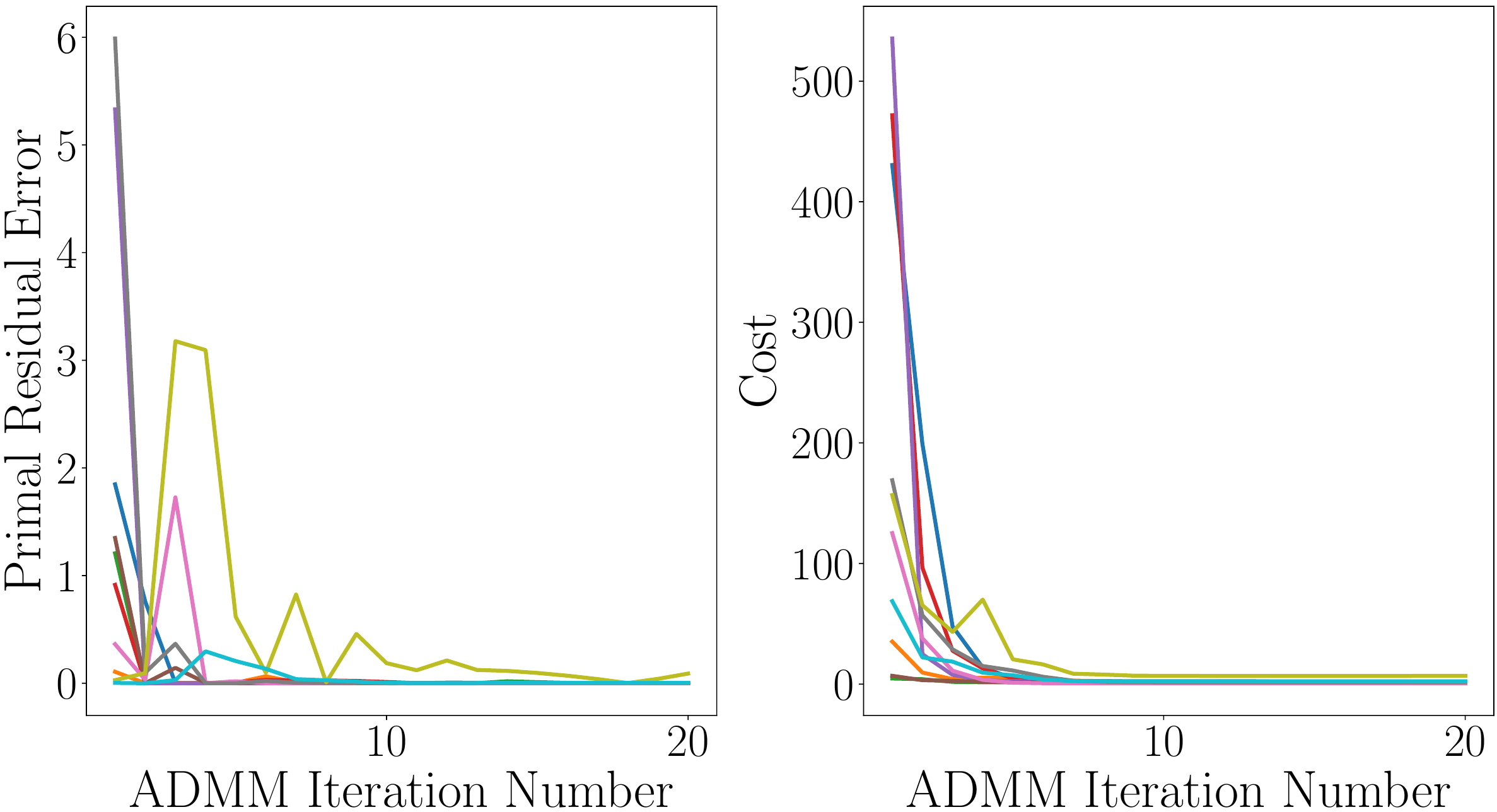}
    \caption{Primal residual error and cost during ADMM iterations of considering directional manipulability in ten simulations.}
    \label{Fig21}
\end{figure}

Based on the above simulation results, we add the consideration of the manipulability in the motion planning problem and design a pick-and-place motion scenario as shown in \autoref{Fig4_1}. We also compare the maximization of different metrics to show the advantage of directional manipulability, as shown in \autoref{Fig4_2}. Here, we treat the tool as an external link in the kinematic chain of the robot after the pick-up stage. In this experiment, the weight of each joint angle actuator is set to be unitary (i.e., with same importance). The algorithm parameter setting is kept the same as in Section \ref{sec:simulation_admm_ilqr_via_range}, except for the control weight of iLQR, which is changed to $1e\textnormal{-}5$. The state and control penalty of ADMM are changed to \{$\bm{Q}_r$-\emph{2D}/\emph{3D}: $\diag(1e\textnormal{-}1)$\}, \{$\bm{R}_r$-\emph{2D}: $\diag(1e\textnormal{-}2)$; \emph{3D}: $\diag(1e\textnormal{-}3)$\}, respectively. The directional manipulability maximization cost weight is set as \{$w_\text{man}$-\emph{2D}: $1e0$; \emph{3D}: $1e\textnormal{-}1$\}. Due to the 3D space simulation task setup close to the actual hammering, its cost function is defined as \autoref{eq:ilqr_cost_final}. The final position reaching weight $w_\text{pos}=1e2$, the weight of the via grasping orientation and final direction reaching are $w_\text{o}=1e1$ and $w_\text{dir}=1e1$. We perform simulations for different methods in ten different experimental settings. The projection value of the velocity manipulability is shown in \autoref{Fig12}. We also plot the primal residual error and cost during ADMM iterations of considering directional manipulability in \autoref{Fig21}. The average computation time for these four different methods is $49.349$s (directional velocity manipulability), $20.014$s (not considering velocity manipulability), $48.607$s (\emph{common metric}), $70.134$s (tracking desired velocity manipulability).

\begin{figure}[h]
	\centering
	\subfigure[Joint angle configurations in the pre-hammering state using different methods to maximize the velocity manipulability along the desired direction.]{	
		\centering
		\includegraphics[width=0.65\columnwidth]{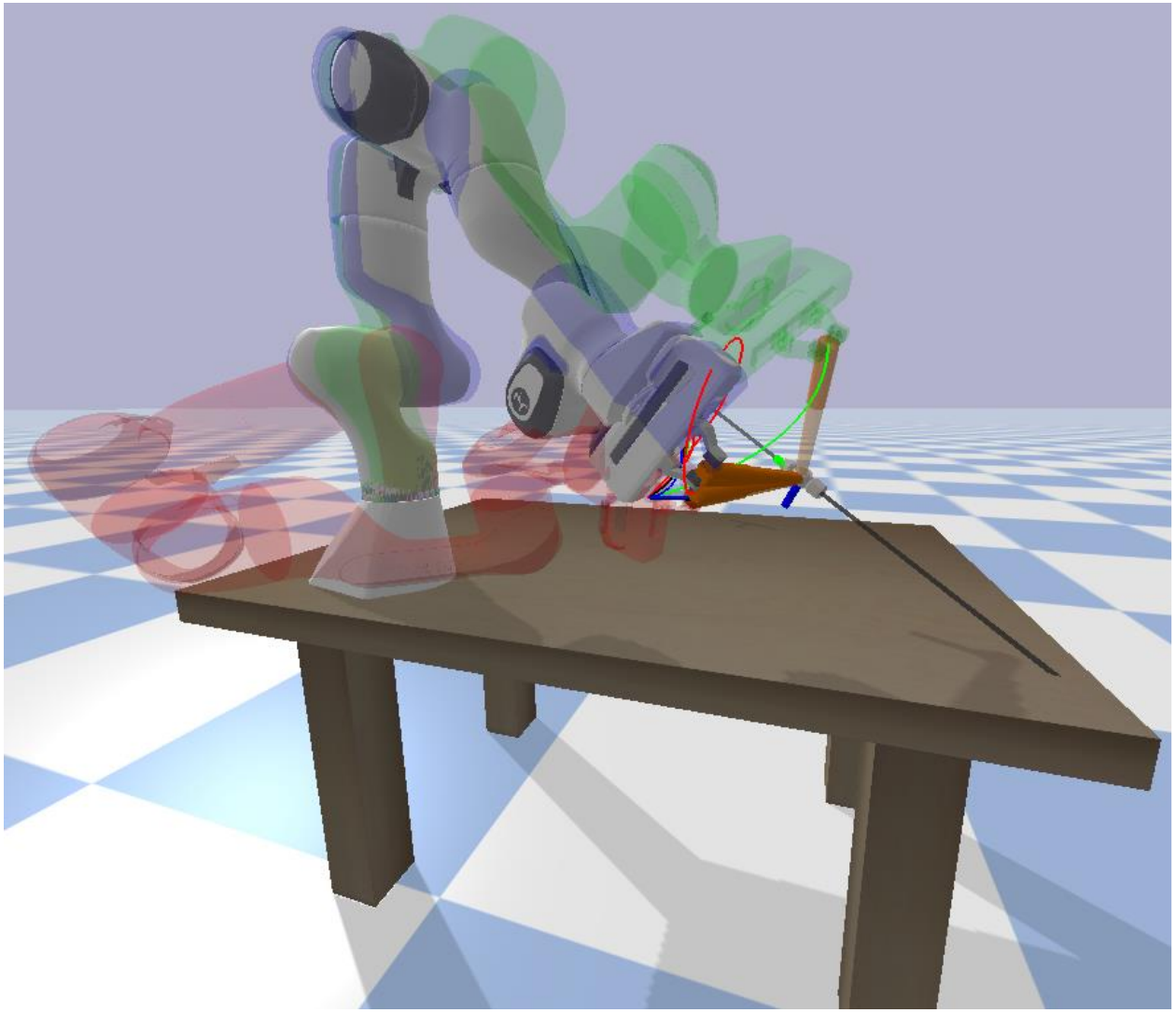}	
		\label{Fig5_1}
	}
	\subfigure[Velocity manipulability visualized in 3D space (left), with its projection along the desired direction (right).]{
		\centering
		\includegraphics[width=\columnwidth]{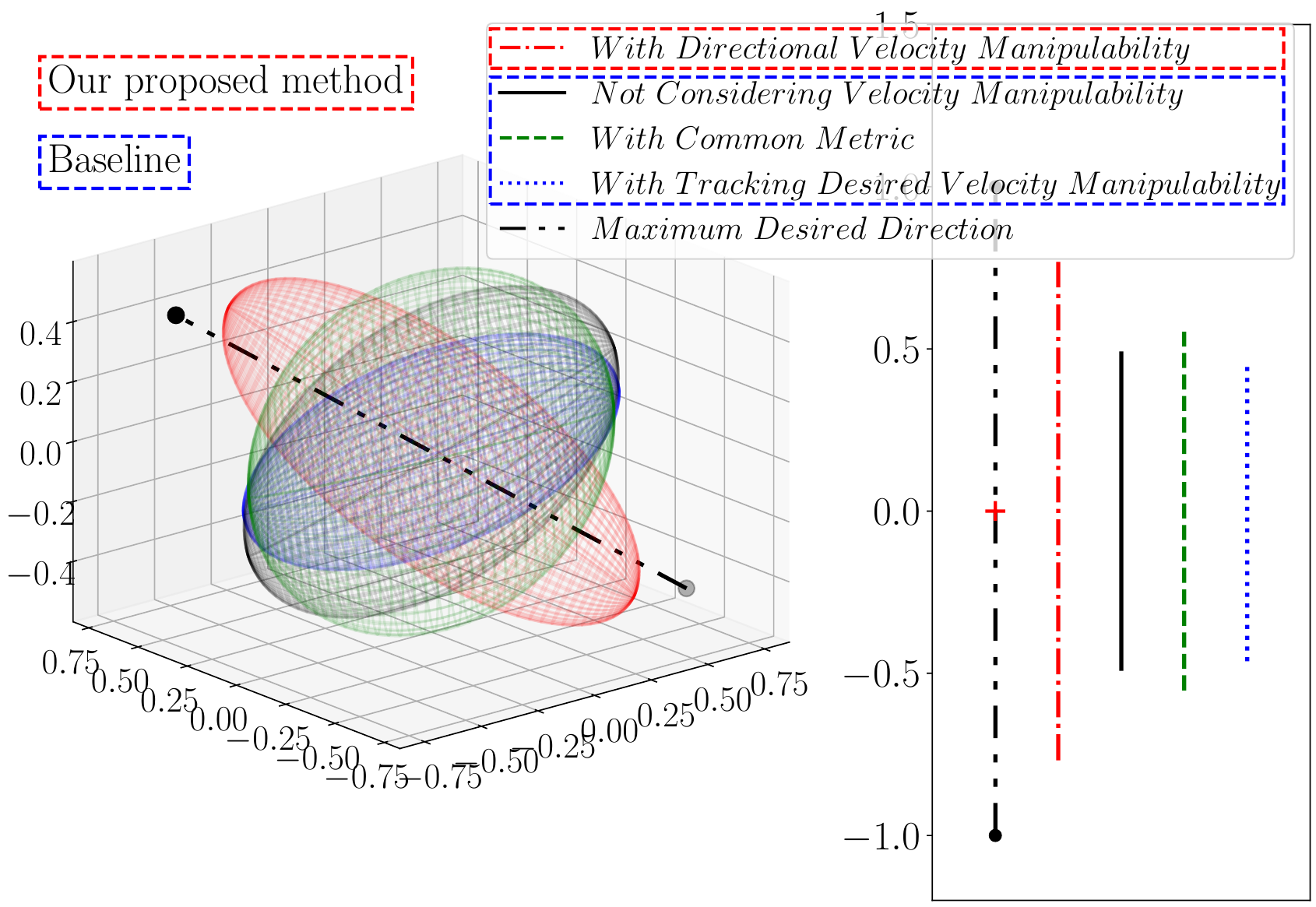}	
		\label{Fig5_2}
	}
	\caption{ADMM-iLQR with maximization of different manipulability metrics in a simulated pick-and-place task. 
 In (a), the red robot corresponds to the use of the directional velocity manipulability. The robot in solid lines does not consider velocity manipulability. The green robot employed the \emph{common metric}. The blue robot tracked a desired velocity manipulability. For a better view of the transparent overlays of the robots, see the color version of this article.}
	\label{Fig5}
	\vspace{-0.3cm}
\end{figure}

We verified the results in a 3D space simulation, as shown in \autoref{Fig5}. We can observe that directional manipulability has a better performance. The method considering directional manipulability can sometimes get stuck in poor local minima, which explains that the volume maximization can sometimes perform better in these cases. The method maximizing the \emph{common metric} enlarges the ellipsoid in an isotropic way. For the approach of tracking a desired manipulability \cite{jaquier2021geometry}, it is hard to design a perfect desired manipulability to track, which may be beyond the robot's physical configuration or put too many constraints on the ellipsoid axes that are not relevant for the task. The average computation time for these four different methods is $161.7$s (directional velocity manipulability), $147.3$s (not considering velocity manipulability), $163.1$s (\emph{common metric}), $210.5$s (tracking desired velocity manipulability). 


\section{Robot Experiment}
\label{sec:robot experiment}
\begin{figure}[tb]
 	\centering
 	\includegraphics[width=\columnwidth]{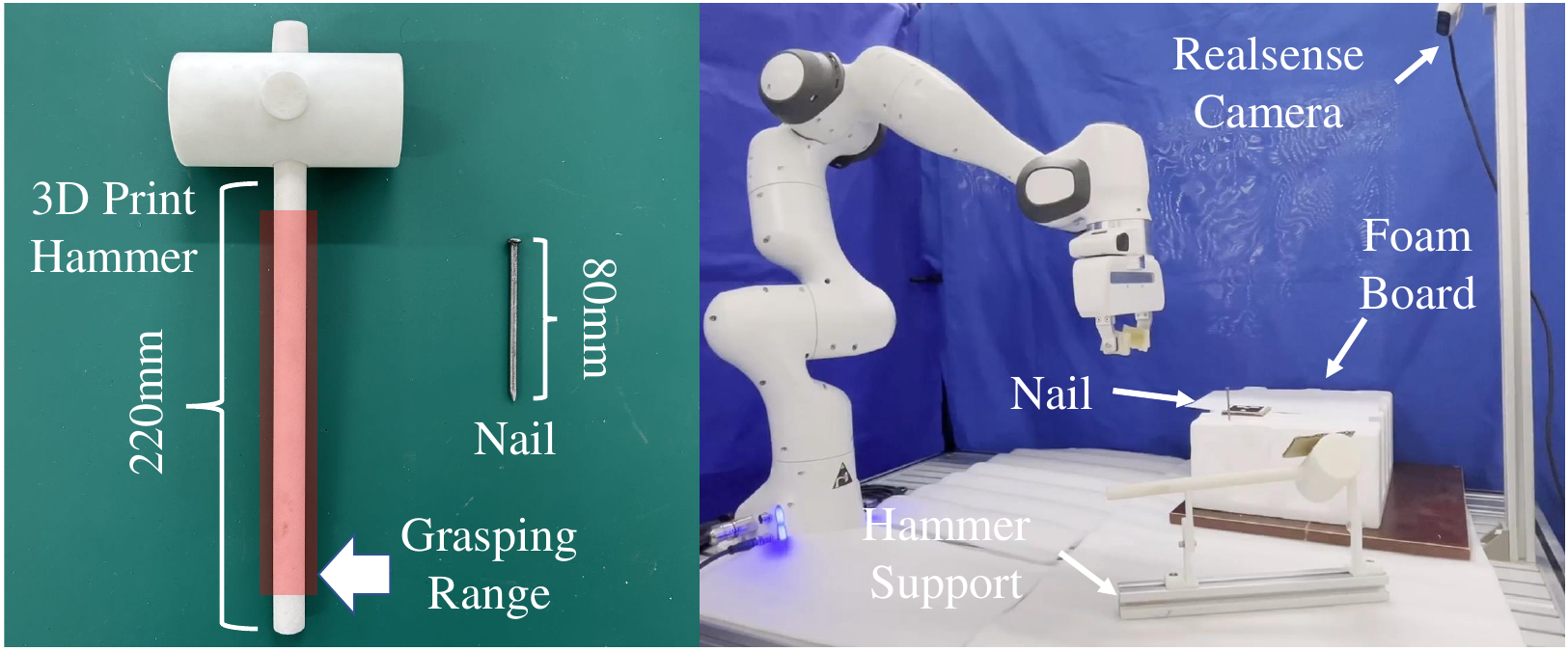}
 	\caption{Experimental setup. The red rectangle represents the available grasping points.}
 	\label{Fig9}
 	\vspace{-0.3cm}
 \end{figure}

We applied our approach to an impact-aware task requiring a robot to hammer a nail into a foamed plastic base. We evaluated the approach with a 7-axis Franka Emika manipulator by using a joint velocity controller on the same PC platform as the simulation. We chose the foamed plastic as the pegboard, which is a soft material with a structure that is still firm inside. This material can mitigate the impulse of the hammering impact on the type of robot we used for the experiment. Then, we used a nail to create a pilot hole by inserting a nail fully and straight into the vacant space of the foam platform. To simulate the connection between the human hand and the hammer, we stuck two pieces of sponge on the two grippers of the robot hand. The hammer is made of ABS material by 3D printing, with the handle designed as a cylinder for easy grasp by a simple two-finger gripper. The grasping range is defined as a cuboid whose length is from 50 mm to 230 mm starting from the head and whose width and height are 1 mm. The setup is depicted in \autoref{Fig9}. The control inequality constraint of the problem is set as the Franka Emika safety joint velocity limit. The time step, trajectory horizon, initial control command, threshold, and other parameters setting of the iLQR algorithm
and ADMM algorithm are the same as those of the 3D space simulation parameters setting in Section \ref{sec:pick_place_mani}. The hammer picking-up time step is set as $t'=T/2$.

\begin{table}[]
    \centering
    \caption{Maximum velocity for each joint of the Franka Emika robot${}^\star$}
    \begin{tabular}{ccc}
    \hline
         Joint index & Joint 1-4 & Joint 5-7  \\
    \hline
         $\dot{q}_{max}$ & 2.1750 rad/s & 2.610 rad/s \\
    \hline
    \end{tabular}
    \begin{tablenotes}
      \small
      \centering
      \item  ${}^\star$https://frankaemika.github.io/docs/control\_parameters.html
    \end{tablenotes}
    \label{tab:my_label}
\end{table}

\subsection{Hammering Force Test}
\begin{figure}[t]
	\centering
	\subfigure[Relation between the maximum impact force and the nail insertion depth]
        {	
		\centering
		\includegraphics[width=0.44\columnwidth]{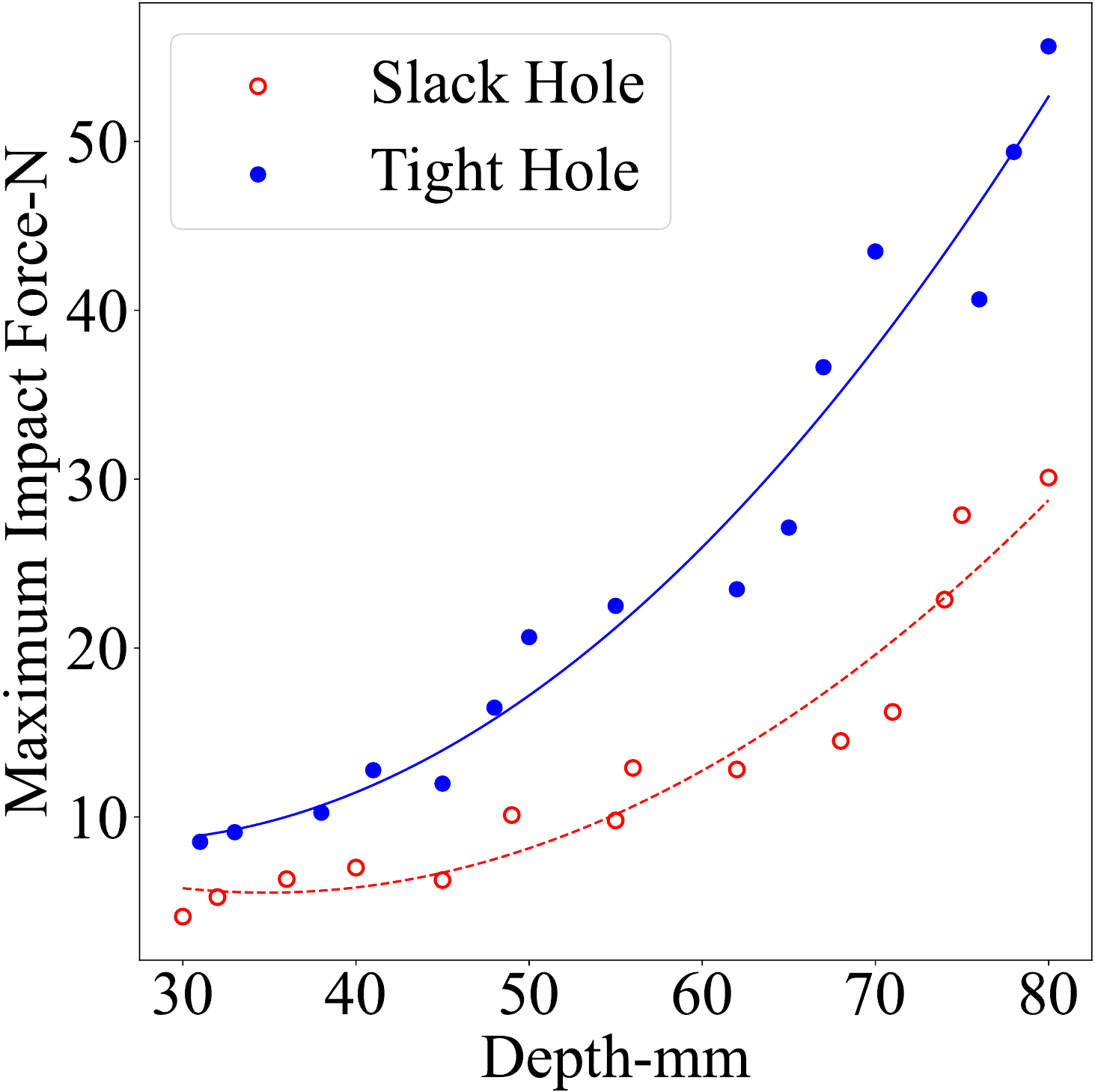}	
		\label{Fig6_1}
	}
        \hspace{4mm}
	\subfigure[Experimental setup for impact force sampling]
        {	
		\centering
		\includegraphics[width=0.43\columnwidth]{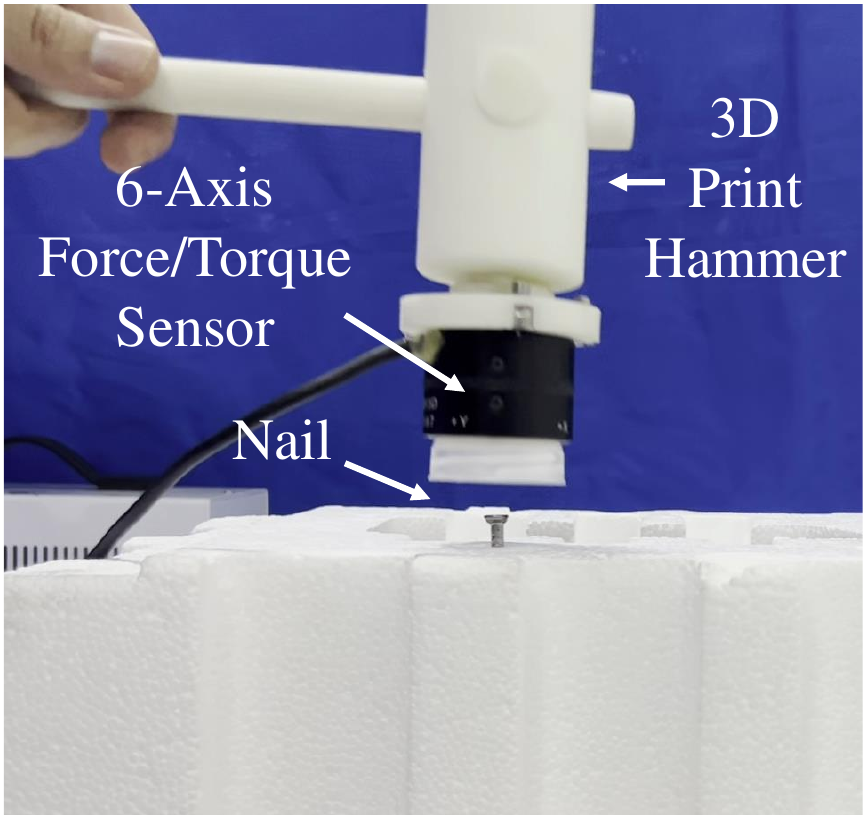}	
		\label{Fig6_2}
	}
	\caption{Relationship between the insertion depth of the nails and the impact force for two levels of tightness. The red dashed line and the blue solid line represent the modeled relation based on the 15 sets of data sampled in the slack and tight cases.}
	\label{Fig6}
	\vspace{-0.3cm}
\end{figure}

We chose two kinds of tightness to evaluate our approach. Due to the elasticity of the hammering platform, we could adjust the tightness by repeatedly inserting and pulling the nail to enlarge the pilot hole. Due to experimental equipment constraints, we could not record impact forces accurately. Indeed, as the connection between the grippers and the tool remains soft, it is hard to fully transmit the impact force to the robot's sensors. To cope with this issue, we modeled the relationship between the insertion depth of the nails and the maximum impact force. We then exploited this model to derive the impact force from the observed insertion depth of the nails. We sampled 15 sets of impact forces with the insertion depth of the nails, for each different tightness state (blue dots represent tight holes, red circles represent slack holes). A quadratic regression model was used to encode this relationship. The sampling data and setup are shown in \autoref{Fig6}. Note here that if an external force sensor were mounted on the head of the hammer, it would increase the weight of the hammer head and cause a bias for the gravity center, further increasing the grip instability and making subsequent impact experiments difficult. 

\subsection{Hammering Nails}
For the real robot experiment, we additionally used the insertion depth of the nails as a way to visualize the differences in velocity manipulability produced by different strategies during impact. We randomly set the position of the hammer and nail within the workspace of the Franka Emika robot and fixed the initial joint angle configuration of the robot. Inspired by the definition of manipulability, we normalized the joint velocity vector, which is calculated by inverse kinematics to keep the hammering direction along the nail axis. We neglected the direction bias during the impact process. The normalized motion in joint space results in a difference in the manipulability ellipsoid on the end-effector. In the experiment, we used a joint velocity controller to drive the system. We kept the hammering velocity of each joint constant during the impact process. In the hitting stage, the robot first drives the hammer back two time steps at strike speed so that there is enough space to reach the target speed at the state of hitting the nail.

\begin{figure*}
	\centering
	\includegraphics[width=0.8\linewidth]{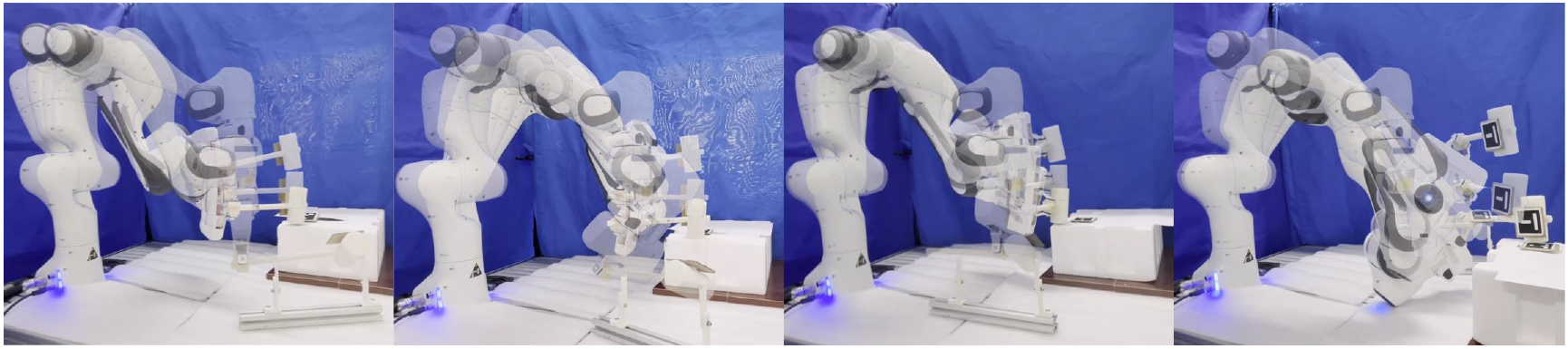}
	\caption{Screenshot of four typical cases using the ADMM-iLQR with maximizing the directional velocity manipulability in the hammering task.}
	\label{Fig8}
	\vspace{-0.3cm}
\end{figure*}

Screenshots of the hammering process are shown in \autoref{Fig8}, depicting four typical cases, including the entire trajectory and the grasping and pre-hammering state in \autoref{Fig10}. Additional experiment details are presented in the supplementary video. Due to the limited workspace available for the experiment, we chose twelve locations of hammer and nail that were reachable to accomplish the entire impact-aware task (six experiments for each manipulability metric, using two different tightness values). We used the insertion depth of the nails as a performance indicator for each method. In addition, we contrasted the use of ADMM-iLQR for motion planning to the use of kinesthetic demonstrations (Human Demonstration baseline). The use of kinesthetic demonstrations is another intuitive and effective way to accomplish manipulation tasks when operating a robotic manipulator. Therefore, we used this as one of the baselines. But unlike demonstrations in point-to-point movements, impact-aware tasks not only require reaching a set of desired viapoints but also ensuring efficient motion of the manipulator and efficient exploitation of the kinematic redundancy to improve manipulability. To realize that, the user dragged the arm of the robot to grasp the hammer and then put the arm into a pre-hammer state.

\begin{table}[hbp]
    \centering
    \caption{The projection values along the desired direction of the twelve times hitting experiment}
    \begin{tabular}{cccccc}
    \hline
         Index & A & B & C & D & E \\
         \hline 
         S1 & 0.5894 & \textbf{0.7567} & 0.6677 & 0.6276 & 0.6084 \\
         S2 & 0.5224 & \textbf{0.7168} & 0.6322 & $\times$ & 0.5057 \\
         S3 & 0.6267 & \textbf{0.7346} & 0.5601 & 0.5158 & 0.5843 \\
         S4 & 0.7443 & \textbf{0.8800} & 0.8285 & 0.8257 & 0.8534 \\
         S5 & 0.6838 & \textbf{0.7700} & 0.6868 & 0.5744 & 0.5934 \\
         S6 & 0.7252 & \textbf{0.7505} & 0.6715 & 0.6305 & 0.6942 \\
         \hline
         T1 & 0.5674 & \textbf{0.7579} & 0.6652 & 0.6223 & 0.6197 \\
         T2 & 0.5970 & \textbf{0.7507} & 0.7395 & 0.6921 & 0.6485 \\
         T3 & 0.6900 & \textbf{0.7962} & 0.6157 & 0.6234 & 0.5456 \\
         T4 & 0.5849 & \textbf{0.7367} & 0.6375 & 0.5875 & 0.6813 \\
         T5 & 0.7367 & \textbf{0.8884} & 0.8207 & 0.8303 & 0.8434 \\
         T6 & 0.4948 & \textbf{0.6227} & 0.5863 & $\times$ & 0.4923 \\
         \hline
    \end{tabular}
    \begin{tablenotes}
      \small
      \item (A) Not considering velocity manipulability; (B) Maximization of directional manipulability; (C) Common metric; (D) Desired ellipsoid tracking method; (E) Human demonstration.
    \end{tablenotes}
    \label{tab:table1}
\end{table}

\begin{table}[htbp]
    \centering
    \caption{The insertion depth of the nails}
    \begin{tabular}{cccccc}
    \hline
         Index & A & B & C & D & E \\
         \hline 
         S1 & 63 & \textbf{75} & 70 & 70 & 71 \\
         S2 & 65 & \textbf{71} & 69 & $\times$ & 63 \\
         S3 & 68 & \textbf{73} & 69 & 60 & 65\\
         S4 & 72 & \textbf{80} & \textbf{80} & 76 & \textbf{80} \\
         S5 & 70 & \textbf{80} & 75 & 68 & 70 \\
         S6 & 71 & 74 & \textbf{75} & 73 & 73 \\
         \hline
         T1 & 39 & \textbf{66} & 55 & 43 & 45 \\
         T2 & 40 & \textbf{65} & \textbf{65} & 64 & 54 \\
         T3 & 45 & \textbf{70} & 48 & 47 & 34 \\
         T4 & 41 & \textbf{61} & 45 & 46 & 63 \\
         T5 & 45 & \textbf{75} & 68 & 68 & 68 \\
         T6 & 35 & \textbf{58} & 42 & $\times$ & 31 \\
         \hline
    \end{tabular}
    \begin{tablenotes}
      \small
      \centering
      \item Indices A/B/C/D/E correspond to the indices in \autoref{tab:table1}; values in $mm$.
    \end{tablenotes}
    \label{tab:table2}
\end{table}

\begin{figure}
	\centering
	\includegraphics[width=\columnwidth]{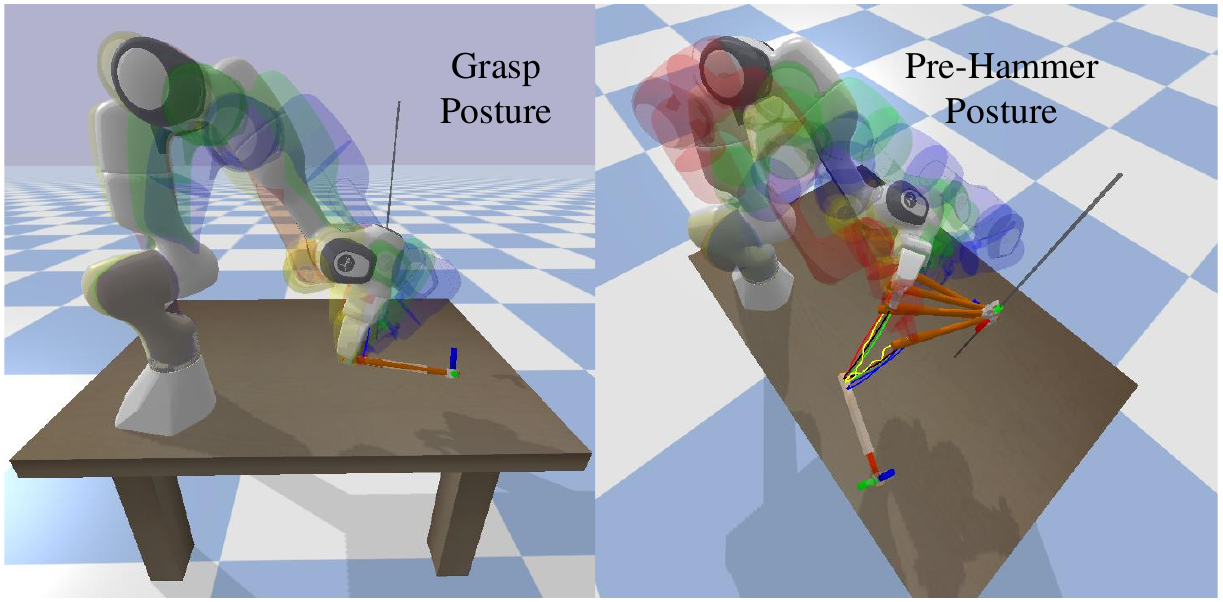}	
	\caption{Trajectories in the case of different maximization strategies. The left and right plot represent the grasping posture and the final pre-hammering state, respectively. The methods for different colored robots are the same as shown in \autoref{Fig5_1}, with the addition of the baseline using human demonstration (yellow robot). For a better view of the transparent overlay of the robots, see the color version of this article.}
	\label{Fig10}
	\vspace{-0.3cm}
\end{figure}

\begin{figure}[h]
	\centering
	\includegraphics[width=\columnwidth]{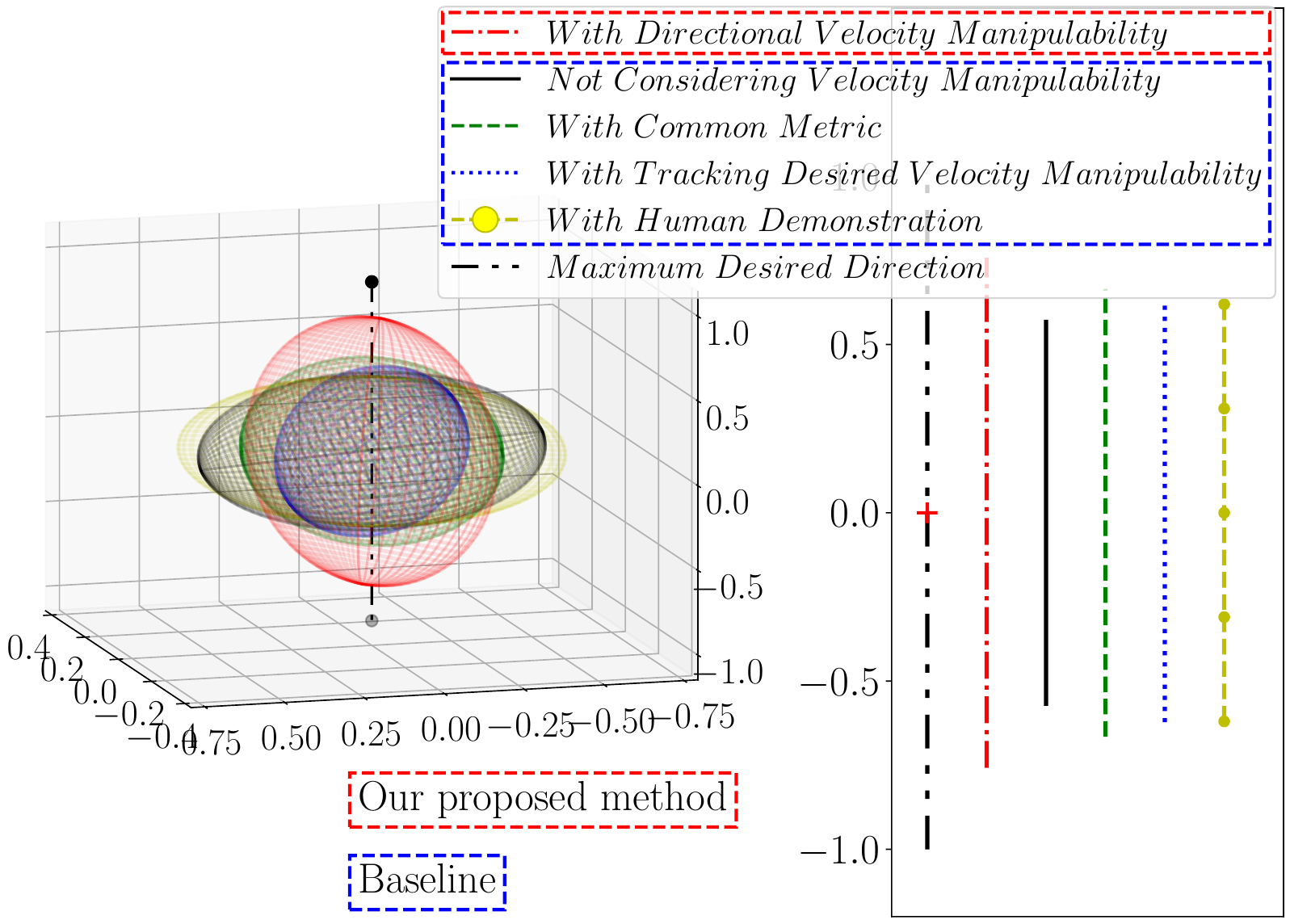}
	\caption{Velocity manipulability of the robot-hammer system at the impact point, generated by different methods. These ellipsoids represent the velocity manipulability of the robot-hammer system, whose end-effector is located at the hammer head. The projection results are displayed on the right part for the method for directional velocity manipulability (in red dash-dotted line) without considering velocity manipulability (in black solid line), for \emph{common metric} velocity manipulability (in green dashed line), for desired velocity manipulability tracking (in blue dotted line), and from human demonstration (in yellow dash-dotted line). For a better view of the transparent overlayed ellipsoids, see the color version of this article.}
	\label{Fig11}
	\vspace{-0.3cm}
\end{figure}
\subsection{Discussion of results}
\begin{figure}[htbp]
    \centering
    \includegraphics[width=0.9\columnwidth]{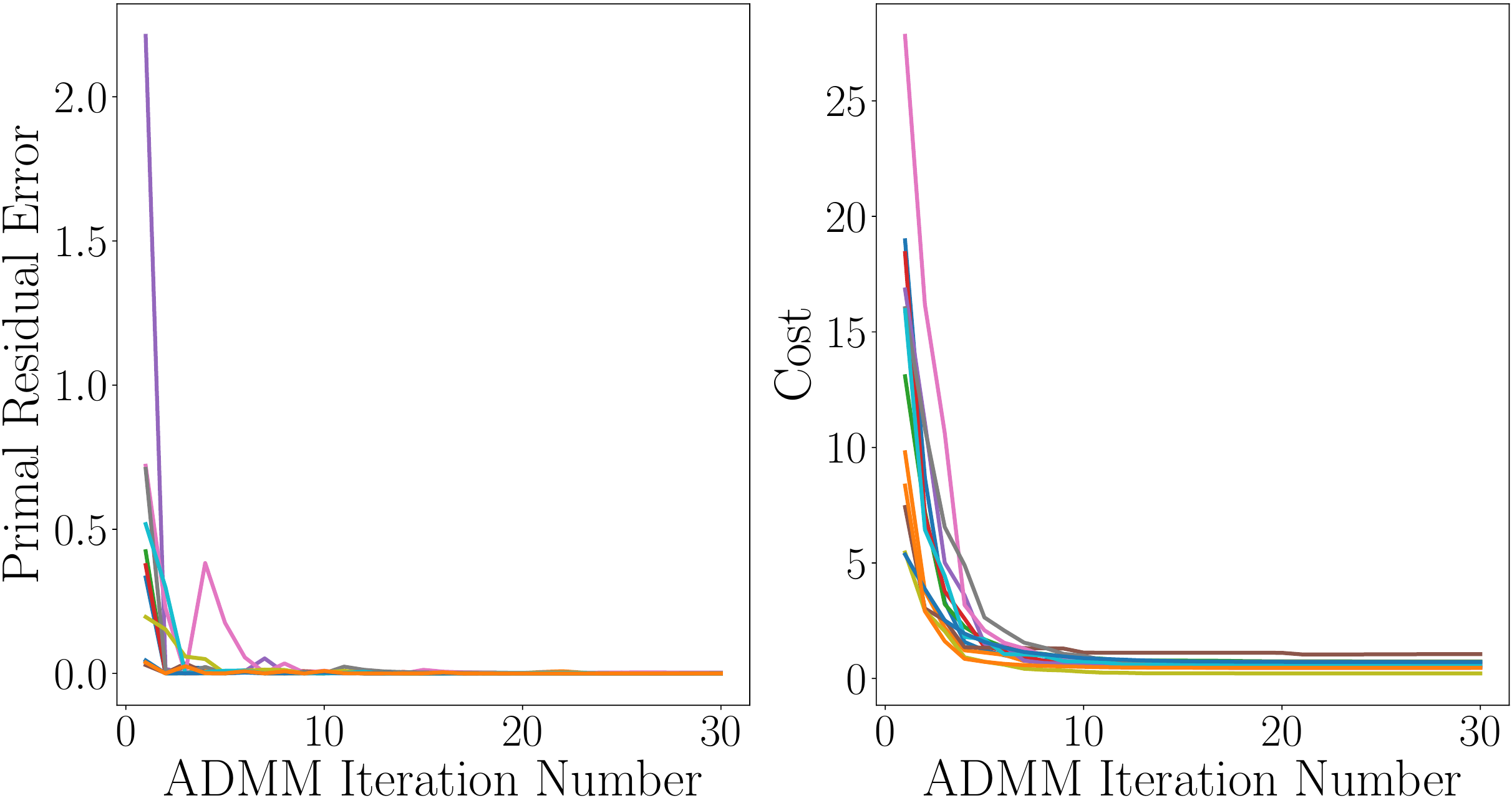}
    \caption{Primal residual error and cost during ADMM iterations when considering directional manipulability in twelve hammering experiments.}
    \label{Fig22}
\end{figure}

\begin{figure}[h]
	\centering
	\includegraphics[width=0.75\columnwidth]{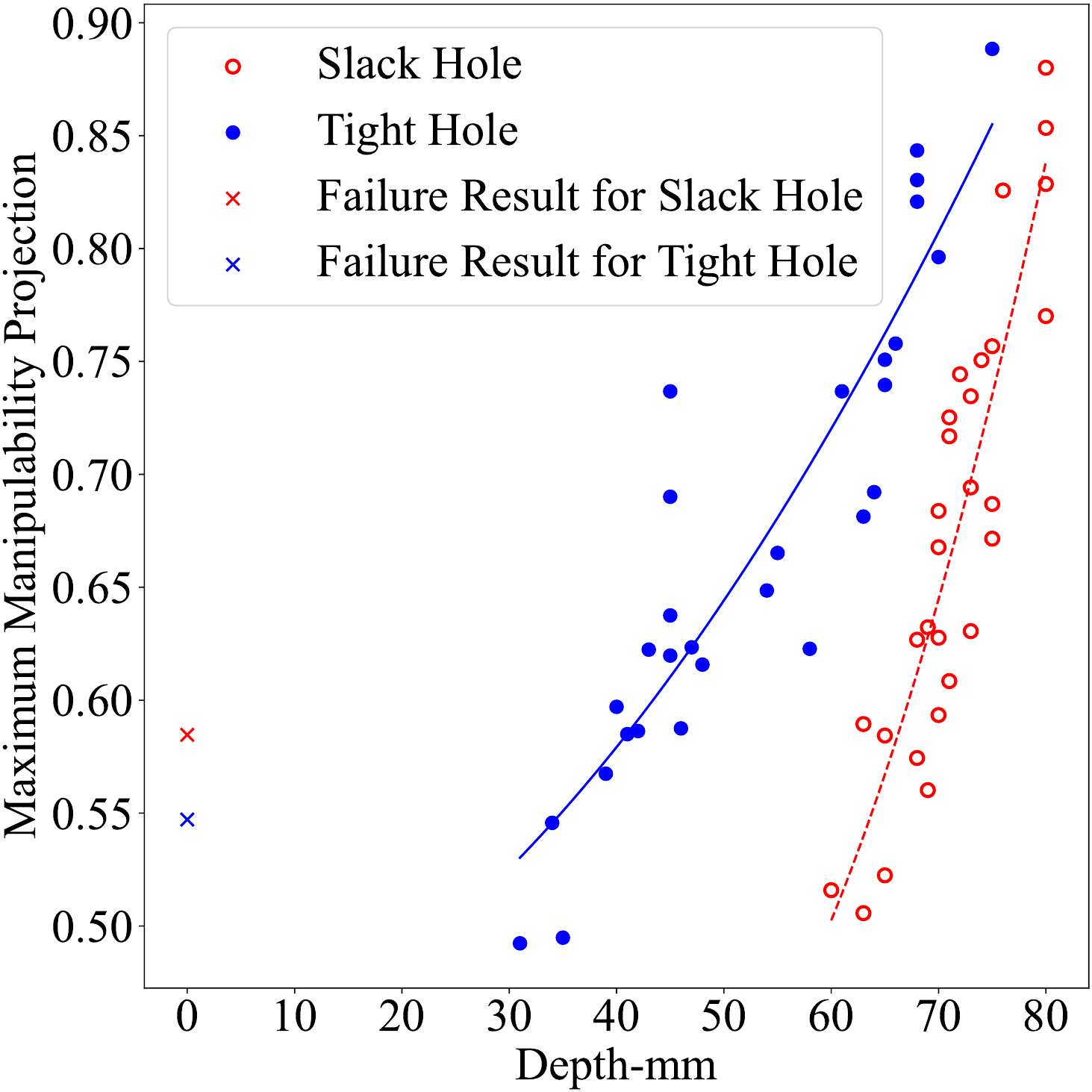}	
	\caption{Relations between the nail insertion depths and the maximum projection of manipulability along the desired direction.}
	\label{Fig20}
	\vspace{-0.3cm}
\end{figure}

\begin{figure}
	\centering
	\includegraphics[width=\columnwidth]{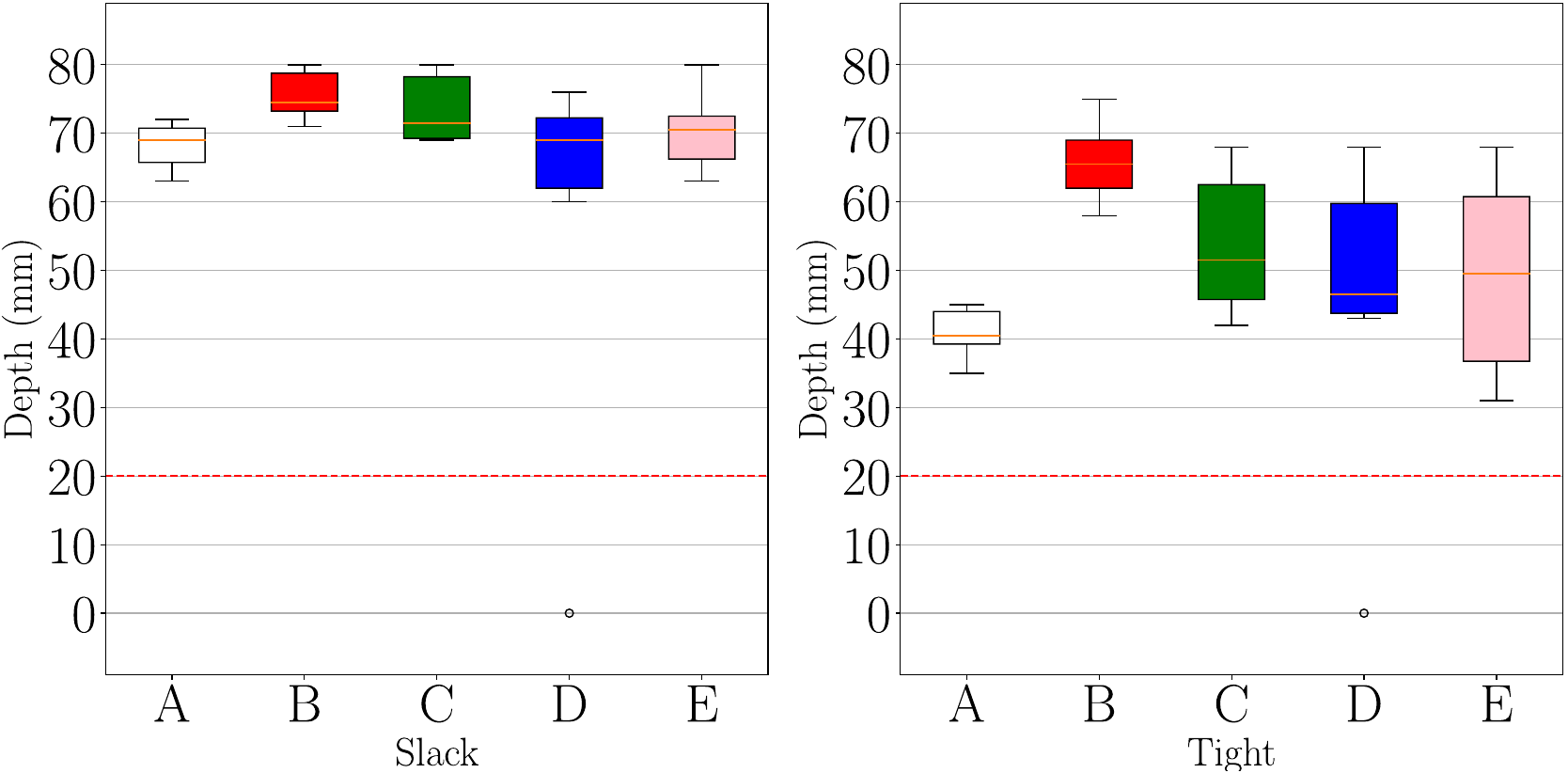}	
	\caption{Nail insertion depths in slack and tight cases. The boxplots show the results for the ADMM-iLQR motion planning using different methods, where the indices A/B/C/D/E correspond to the indices in \autoref{tab:table1}. Circle points correspond to the cases where the robot finds solutions over the joint limits in the motion planning process. The red dashed line represents the pre-depth of the nail in the plastic foam.}
	\label{Fig7}
	\vspace{-0.3cm}
\end{figure}

We list the maximum projection along the desired direction in \autoref{tab:table1} and the corresponding insertion depth of the nails in \autoref{tab:table2}. We can observe that the optimization considering the directional manipulability has the highest value in the twelve experiments. However, in the depth case S4, B/C/E all have the same performance because the impact force has exceeded the maximum limit of definable depths. In \autoref{tab:table2} case S6, these methods all perform similarly, but C has the highest value. Indeed, although the pilot hole smooths out the hammering task, the rotational and plastic deformation during hammering remains, which could still cause in some cases the nail to be rotated by the hammering deflection. It is also the reason for the inconsistency in the relationship between the projection values and the insertion depth of the nails for the different methods. We selected one case from the four typical cases in \autoref{Fig8} to illustrate the entire trajectory and plot the grasping tool's gestures and the pre-hammering robot's configuration in \autoref{Fig10}. We plot the velocity manipulability of the robot in its hammering pose, and its projection along the desired direction of the hammering on the right, where we can find that the projection of the directional manipulability is the largest. The primal residual error and cost during ADMM iterations of considering directional manipulability are shown in \autoref{Fig22}. We also plot the relation between the projection and the insertion depth in \autoref{Fig20}, where we can find that its trend is similar to \autoref{Fig6_1}. The gradient of the maximum impact force increases gradually as the insertion depth increases. To observe the performance of the different methods, the box plot in \autoref{Fig7} additionally compares the insertion depth of the nails using different manipulability metrics. We can observe that almost all the methods perform well in the slack nail driving case. In the tight case, the human demonstration method is randomly distributed due to the robot configuration not being considered, where the user typically prefers to use an easily dragged posture in the demonstration. The directional manipulability performs best, and the \emph{tracking desired ellipsoid} strategy does not work in some task settings. The \emph{common metric} strategy focuses on volume maximization, which only worked efficiently in some cases. The average computation time for these four different methods is $56.8$s (directional velocity manipulability), $44.1$s (not considering velocity manipulability), $50.7$s (common metric), $97.5$s (tracking desired velocity manipulability). There is a big difference between the actual scene and the simulation on the hammer's placement and the hitting direction setup, significantly impacting the computation time. In the simulation, the hammer and the hitting direction are randomly set in the table workspace. However, in the actual task, the hitting direction should be set perpendicular to the ground at a specific height. Those result in a significant difference in the computation time of both cases. Since the motion planning of the final hitting motion is not considered in the optimization process, the failures due to reaching the joint limits (listed in the table) all occurred during the impact phase. We can observe that tracking the desired manipulability always leads the robot to singular configurations close to joint angle limits, which makes it easy for the impact motion to exceed the joint physical limits (see hollow ring of the blue box in \autoref{Fig7}). This kind of configuration has a significant influence on the subsequent hammering action. Therefore, we defined this situation as a failure case.

In addition to the above successful cases, we also encountered some failure cases that were not considered in the analysis of the results. In the hammer grasping phase, due to the limitations of the experimental equipment, we employed a simple gripper. When handling the hammer, the gripper cannot guarantee a very stable grasp because the gravity is mainly distributed on the head of the hammer. Therefore, the held hammer can sometimes be rotated or shifted when grasping it or in the subsequent movement process. In addition, when the robot picks up the hammer away from the support frame, the grasp can also be affected by perturbations caused by the support frame. In the proposed approach, the tool handled by the robot is considered as an external link of the robot kinematic, so the above disturbances can degrade the reaching of the final hitting pose. Also, as our  contribution focused on the tool affordance aspects, we did not consider the collision avoidance problem extensively. While collision is an additional possible failure factor, it rarely happened in practice with the problem setup that we proposed, except in the comparison case of tracking a desired manipulability.

\section{Discussion}
\label{sec:discussion}
The approach presented in this paper adopts the ADMM-iLQR approach to solve motion planning problems exploiting tool affordances by comparing different strategies for maximizing manipulability along specific task directions with a constrained optimal control approach. The gripping point for the tool is treated as a viapoint to reach within a desired range, which provides the flexibility of choosing the grasping posture, so that the robot can automatically select an ideal gripping point by anticipating how the tool will be further used after grasping it. Thus, the grasping hand pose and the body configuration are also determined by the final hitting direction, allowing the motion planning process to use the tool by considering the requirements of both the task and the environment in which the motion takes place. By considering the directional velocity manipulability, the manipulator can give a large local speed to the handheld tool, effectively generating a large momentum during the impact. 

In contrast to the scenario where a robot hand would directly act on the nail, using a tool such as a hammer involves an additional soft connection in the kinematic chain, where the tool can be used as an additional link in the kinematic chain. The soft connection leads to a quasi isolation of the body inertia from the system during the impact, where the tool inertia has the most important influence. The assumption for this category of hitting tasks using tools is that the robot inertia can be ignored during the impact. In contrast, the robot posture will influence the capability of the robot to use the handheld tool. Velocity manipulability is thus maximized rather than inertial or force manipulability. 

In the experiment, we used a simple two-finger gripper that cannot ensure perfectly stable grasping and hammering. Although the two-finger can meet our simple requirement, we still sporadically observed unexpected outcomes, such as the hammer falling off during the impact motion, or a deviation in the hitting impact. In future work, we aim to ameliorate this aspect by combining the approach with a visual corrective feedback component.

\section{Conclusion}
\label{sec:conclusion}
In this article, we proposed a constrained optimal control approach considering directional manipulability as a way to improve the performance of using tools in impact-aware tasks. We relied on ADMM-iLQR to efficiently solve constrained optimal control problems, by considering the notion of tool affordance, in the form of task space constraints in which the robot can determine an optimal configuration to grasp the tool by taking into account what the tool can offer when held by the robot. The approach has been tested for tasks requiring impacts. We demonstrated in various simulations and on a real robot experiment that the robot was able to solve this challenge with viapoint references specified as a desired range to reach with preferred orientations, while maximizing the velocity manipulability along desired task directions. Our experimental results showed that the consideration of directional manipulability in our approach achieved better hammering performance than the baselines, including the use of human demonstrations, the use of a common scalar manipulability index, and the tracking of desired manipulability ellipsoids.


\section*{Acknowledgments}
This work was funded by the National Key Research and Development Program of China (No. 2022YFB4700701) and by the Major Research Plan of the National Natural Science Foundation of China (No.92048301), supported by the China Scholarship Council (CSC, No.202006120159). The authors would like to thank Dr Hakan Girgin for his suggestions and support on the use of ADMM.

{\appendix}
{\appendices}


\section*{Projection onto sublevel set of a convex function}
Solutions to the optimization problems in the form $\underset{\bm{x}}{\min} \|\bm{x}-\bm{z}\| \; \text { s.t. } \; f(\bm{x}) \leq u$ is called a projection onto the sublevel set of a convex function and denoted as $\Pi_{\mathcal{C}}(\bm{z})$ with the convex set $\mathcal{C}=\{\bm{x} \mid f(\bm{x})-u \leq 0\}$. This projection is
given by $\Pi_{\mathcal{C}}(\bm{z})=(\bm{I}+\mu \bm{\partial} f)^{-1}\bm{z}$, where $\mu$ is an arbitrary solution of $f\left(\Pi_{\mathcal{C}}(\bm{z})\right)=t$.

\noindent\textbf{Affine projections:\quad} If we let $f(\bm{x})=\bm{a}^{\trsp} \bm{x}$ with $\mathcal{C}=\left\{\bm{x} \mid \bm{a}^{\top} \bm{x} \leq u\right\}$, then $\bm{\partial} f=\bm{a}$, hence
\begin{equation}
	\begin{aligned}
		(\bm{I}+\mu \bm{\partial} f) \Pi_{\mathcal{C}}(\bm{z})=\bm{z}, \\
		\Pi_{\mathcal{C}}(\bm{z}) + \mu \bm{a} = \bm{z}, \\
		\Pi_{\mathcal{C}}(\bm{z}) = \bm{z} - \mu \bm{a}.
	\end{aligned}
\end{equation}

Then, we find an arbitrary solution for $\mu$ from $f(\Pi_{\mathcal{C}}(\bm{z}))=u$ as  
\begin{equation}
	\begin{aligned}
		f(\bm{z}-\mu \bm{a})=u, \\
		\bm{a}^{\trsp}(\bm{z}-\mu \bm{a})=u, \\
		\mu = \frac{\bm{a}^{\trsp} \bm{z} - u} {{\Vert \bm{a} \Vert}^2_2}.
	\end{aligned}
\end{equation}

One can find the projection onto $\mathcal{C}=\{\bm{x} \mid l \leq \bm{a}^{\trsp}\bm{x}\}$ by letting $f(\bm{x})=-\bm{a}^{\trsp}\bm{x}$ and by replacing $u$ with $-l$.

Note that if we take $\bm{x}$ to be one-dimensional and $a=1$, then the problem becomes the projection onto bounds $\mathcal{C}=\left\{x \mid x \leq u\right\}$, which can be solved with the \emph{clipping} operator defined by $\Pi_{\mathcal{C}}(z)=\min (z, u)$. This can be extended to lower bounds with $\Pi_{\mathcal{C}}(z)=\max (\min (z, u), l)$. The projection of affine hyperplane is defined as
\begin{equation}
    \begin{array}{rl}
        &\text{Constraints: } l\leq \bm{a}^\trsp \bm{x} \leq u \\
        &\text{Projection: } \Pi_{\mathcal{C}} =\\
        & \left\{
        \begin{array}{lr}
            \bm{x} \quad&\text{if}\quad l\leq \bm{a}^\trsp\bm{x} \leq u,\\
            \bm{x}-(\bm{a}^\trsp\bm{x}-u)\frac{\bm{a}}{\norm{\bm{a}}^2_2}  &\text{if}\quad \bm{a}^\trsp\bm{x}>u,\\
            \bm{x}-(\bm{a}^\trsp\bm{x}-l)\frac{\bm{a}}{\norm{\bm{a}}^2_2}  &\text{if}\quad \bm{a}^\trsp\bm{x}<l. 
        \end{array}
        \right.
    \end{array}
\end{equation}

\section*{Batch solution of Linear Quadratic Regulator (LQR)}
\label{app:batch_ilqr}
We can describe all states $\bm{x}$ in a trajectory, concatenated in a big vector, as a linear function of the initial state $\bm{x}_1$ and the applied control commands $\bm{u}$ as 
\begin{equation}
	\bm{x} = \bm{S_x}\bm{x}_1+\bm{S_u}\bm{u},
	\label{eq:batch_sys}
\end{equation}
where 
\begin{equation}
	\bm{S_x} = 	\left[
								\begin{matrix}
									I \\
									\bm{A} \\
									\bm{A}^2 \\
									\vdots \\
									\bm{A}^{T-1}
								\end{matrix}
								\right],
	\bm{S_u} = 	\left[
								\begin{matrix}
									\textbf{0} & \textbf{0} & \cdots & \textbf{0}\\
									\bm{B} & \textbf{0} & \cdots & \textbf{0}\\
									\bm{A}\bm{B} & \bm{B} & \cdots & \textbf{0}\\
									\vdots & \vdots & \ddots & \vdots\\
									\bm{A}^{T-2}\bm{B} & \bm{A}^{T-3}\bm{B} & \cdots & \bm{B}
								\end{matrix}
								\right].
	\label{eq:Sx_Su}
\end{equation}
To track a desired reference $\bm{x}_d$, the cost function can be defined as 
\begin{equation}
	c(\bm{x}, \bm{u}) =  (\bm{x}-\bm{x}_d)^\trsp \bm{Q} (\bm{x}-\bm{x}_d) + \bm{u}^\trsp \bm{R} \bm{u},
	\label{eq:batch_cost}
\end{equation}
where $\bm{Q}=\text{blockdiag}\left(\bm{Q}_1, \bm{Q}_2,\cdots, \bm{Q}_T\right)$ and $\bm{R}=\text{blockdiag}\left(\bm{R}_1, \bm{R}_2,\cdots, \bm{R}_{T-1}\right)$. Substituting (\ref{eq:batch_sys}) and (\ref{eq:Sx_Su}) to (\ref{eq:batch_cost}), we obtain
\begin{equation}
    \begin{aligned}
  	c(\bm{x}, \bm{u}) =&\bm{u}^\trsp(\bm{S_u}^\trsp\bm{Q}\bm{S_u}+\bm{R})\bm{u}+2\bm{u}^\trsp\bm{S_u}^\trsp\bm{Q}(\bm{S_x}\bm{x}_1-\bm{x}_d)\\
    &+(\bm{S_x}\bm{x}_1-\bm{x}_d)^\trsp\bm{Q}(\bm{S_x}\bm{x}_1-\bm{x}_d).  
    \end{aligned}
	\label{eq:cost_mix}
\end{equation}

The optimal control command $\bm{u}$ is computed by differentiating (\ref{eq:cost_mix}) with respect to $\bm{u}$ and equating to zero, providing the batch form solution
\begin{equation}
	\bm{u} = \big({\bm{S}_{\bm{u}}^\trsp \bm{Q} \bm{S}_{\bm{u}} + \bm{R}\big)}^{-1}
	\bm{S}_{\bm{u}}^\trsp \bm{Q} (\bm{x}_d-\bm{S}_{\bm{x}} \bm{x}_1).
	\label{eq:u_batch}
\end{equation}

\section*{$\mathcal{S}^d$ manifold}
The exponential and logarithmic maps corresponding to the distance
\begin{equation}
    d(\bm{x},\bm{y})=\arccos(\bm{x}^\trsp\bm{y}),
\end{equation}
with $\bm{x}, \bm{y} \in \mathcal{S}^d$ can be computed as (see also \cite{absil2008optimization})
\begin{equation}
    \bm{y}=\text{Exp}_{\bm{x}}(\bm{u})=\bm{x}\cos(\norm{\bm{u}})+\frac{\bm{u}}{\norm{\bm{u}}}\sin(\norm{\bm{u}}),
\end{equation}
\begin{equation}
    \bm{u}=\text{Log}_{\bm{x}}(\bm{y})=d(\bm{x},\bm{y})\frac{\bm{y}-\bm{x}^\trsp\bm{y}\bm{x}}{\norm{\bm{y}-\bm{x}^\trsp\bm{y}\bm{x}}}.
\end{equation}
\bibliographystyle{IEEEtran}
\bibliography{bib_admm-iLQR}

\vspace{40mm}
\begin{IEEEbiography}[{\includegraphics[width=1in,height=1.25in,clip,keepaspectratio]{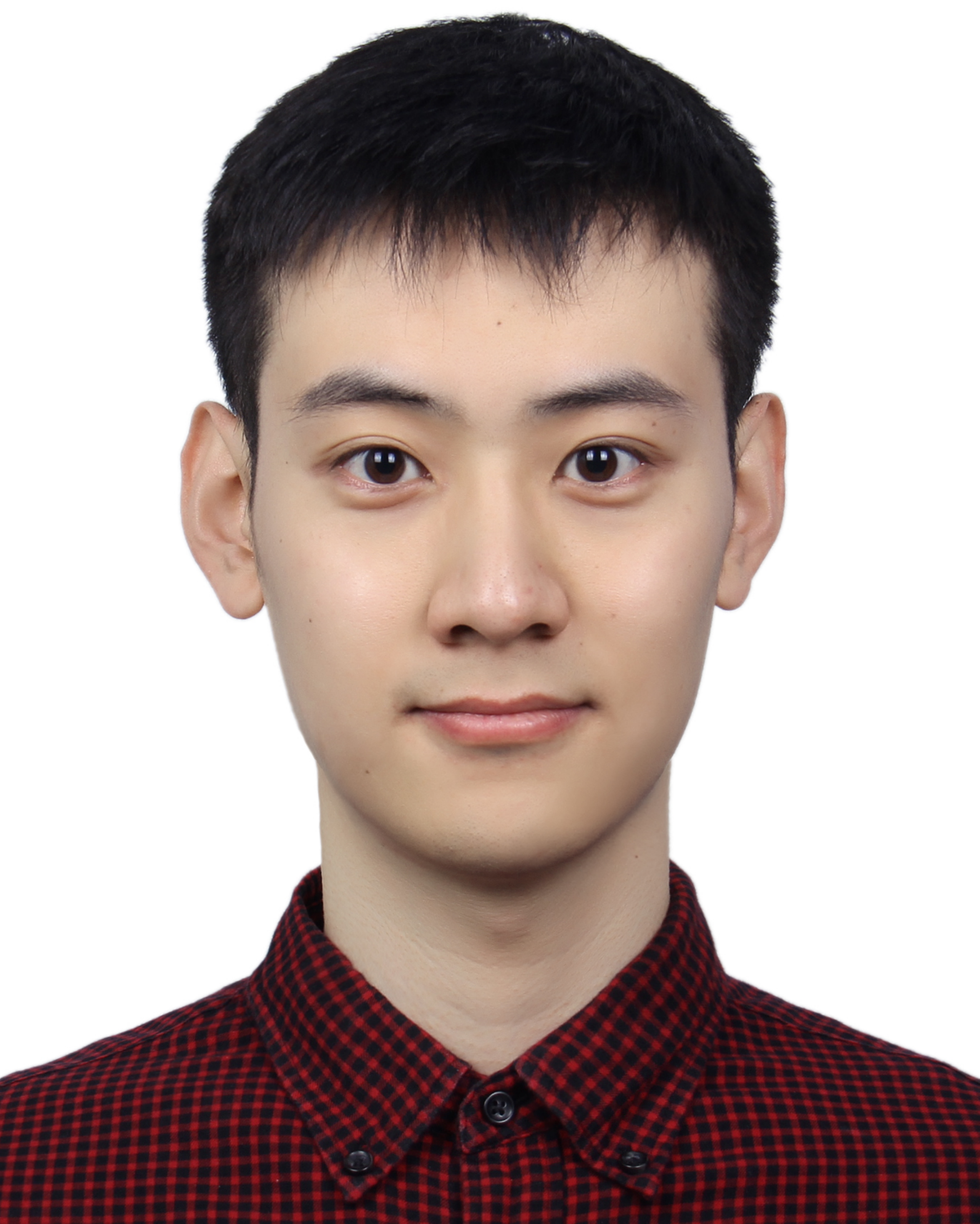}}]{Boyang Ti} received the B.S. degree in mechanical engineering from the Dalian University of Technology (DUT), Dalian, China, in 2017. From 2021 to 2022, he was an intern at the Robot Learning \& Interaction group of Idiap Research Institute. He is currently pursuing the Ph.D. degree with the State Key Laboratory of Robotics and Systems, Harbin Institute of Technology (HIT). His research interests include human-robot collaboration, learning from demonstration, robotic skill learning, and optimal control. Website: https://tflqw.github.io
\end{IEEEbiography}
\vspace{-80mm}
\begin{IEEEbiography}  
[{\includegraphics[width=1in,height=1.25in,clip,keepaspectratio]{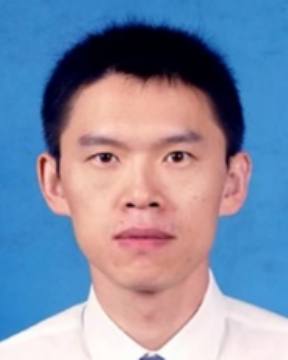}}]{Yongsheng Gao} received the B.Sc., M.Sc., and Ph.D. degrees from the State Key Laboratory of Robotics Institute, Harbin Institute of Technology (HIT), Harbin, China, in 1994, 2001, and 2007, respectively, where he is currently an Associate Professor. His research interests include pathological tremor suppress, tele-operation robot, and biomedical signal processing.
\end{IEEEbiography}
\vspace{-80mm}
\begin{IEEEbiography}  
[{\includegraphics[width=1in,height=1.25in,clip,keepaspectratio]{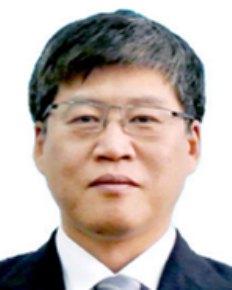}}]{Jie Zhao} received the B.S. and Ph.D. degrees in mechatronics engineering from the Harbin Institute of Technology (HIT), Harbin, China, in 1990 and 1996, respectively. He is currently a Professor with the School of Mechatronics Engineering, HIT, where he is also the Director of the State Key Laboratory of Robotics and Systems. He is the Leader of the Subject Matter Expert Group of Intelligent Robots in the National 863 Program supervised by the Ministry of Science and Technology of China. His research interests include industrial robots and bionic robots.
\end{IEEEbiography}
\vspace{-80mm}
\begin{IEEEbiography}  
[{\includegraphics[width=1in,height=1.25in,clip,keepaspectratio]{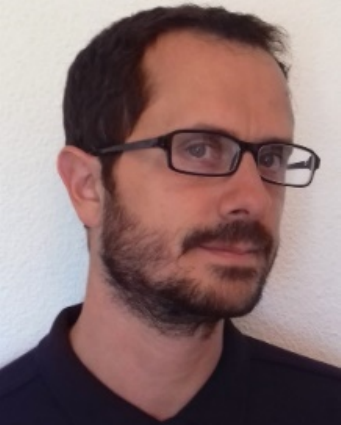}}]{Sylvain Calinon} received the BSc, MSc, and PhD degrees from Ecole Polytechnique Fédérale de Lausanne (EPFL) in 2001, 2003 and 2007, respectively. He is currently a Senior Research Scientist at the Idiap Research Institute and a Lecturer at EPFL. From 2009 to 2014, he was a Team Leader at the Italian Institute of Technology. From 2007 to 2009, he was a Postdoc at EPFL. His research interests cover robot learning, human-robot collaboration, optimal control and model-based optimization. Website: https://calinon.ch
\end{IEEEbiography}
\end{document}